\pdfoutput=1

\documentclass[11pt]{article}

\usepackage[preprint]{acl}

\usepackage{times}
\usepackage{algorithm}
\usepackage{algpseudocode}
\usepackage{amsmath}
\usepackage{mathtools}
\usepackage{amsthm}
\usepackage{amsfonts}
\usepackage{latexsym}
\usepackage{bbm}
\usepackage{multirow}
\usepackage{subcaption}
\usepackage{comment}
\usepackage{setspace}
\usepackage{booktabs}
\usepackage[T1]{fontenc}

\usepackage[utf8]{inputenc}

\usepackage{microtype}

\usepackage{inconsolata}
\usepackage{makecell}

\usepackage{graphicx}

\theoremstyle{plain}
\newtheorem{theorem}{Theorem}[section]
\newtheorem{proposition}[theorem]{Proposition}

\theoremstyle{definition}

\theoremstyle{remark}

\usepackage[normalem]{ulem}
\usepackage{amssymb}%
\usepackage{pifont}%
\newcommand{\cmark}{\ding{51}}%
\newcommand{\xmark}{\ding{55}}%
\useunder{\uline}{\ul}{}
\usepackage{pifont}

\title{Hallucination Detection in LLMs with Topological Divergence on Attention Graphs}

\author{
 \textbf{Alexandra Bazarova\textsuperscript{1}},
\textbf{Andrei Volodichev\textsuperscript{1}},
 \textbf{Aleksandr Yugay\textsuperscript{1}},
 \textbf{Andrey Shulga\textsuperscript{1}},
 \textbf{Alina Ermilova\textsuperscript{1}},
\\
 \textbf{Konstantin Polev\textsuperscript{2}},
 \textbf{Julia Belikova\textsuperscript{2}},
 \textbf{Rauf Parchiev \textsuperscript{2}},
\\
 \textbf{Dmitry Simakov\textsuperscript{2}},
 \textbf{Maxim Savchenko\textsuperscript{2}},
 \textbf{Andrey Savchenko\textsuperscript{2,3}}, 
 \\
 \textbf{Serguei Barannikov$^*$\textsuperscript{1, 4}},
 \textbf{Alexey Zaytsev$^*$\textsuperscript{1}},
\\
\\
 \textsuperscript{1}Applied AI Institute,  \textsuperscript{2}SB AI Lab,\textsuperscript{3} HSE University, 
  \textsuperscript{4}CNRS, Universite Paris Cite
\\
  \small{    \textbf{Correspondence:} \href{mailto:email@domain}{bazarovaai.239@gmail.com}
 }
}

\begin{document}
\maketitle

\def\thefootnote{*}\footnotetext{Equal contribution.}
\begin{abstract}
Hallucinations remain a critical challenge for large language models (LLMs), particularly in Retrieval-Augmented Generation (RAG) settings where models may generate outputs unsupported by the provided context. To address this, we introduce TOHA, a TOpology-based HAllucination detector, which leverages a topological divergence metric to quantify the structural properties of graphs induced by attention matrices. Examining the topological divergence between prompt and response subgraphs in RAG settings reveals consistent patterns: higher divergence values in specific attention heads correlate with unfaithful outputs, independent of the dataset. Extensive experiments — including evaluations on question-answering and summarization tasks — show that our approach achieves state-of-the-art or competitive results across several benchmarks while requiring minimal annotated data and computational resources. Our findings indicate that the topological structure of attention matrices provides an efficient and robust metric for assessing the correctness of LLM's responses. Our source code is publicly available: 
\url{https://github.com/sb-ai-lab/TOHA}. 
\end{abstract}
\section{Introduction}

Large language models (LLMs) have advanced significantly in recent years, with applications across various fields~\cite{chkirbene2024large}. 
To ensure factual reliability, modern LLMs are often combined with the retrieval-augmented generation (RAG) technique, which integrates relevant external knowledge from diverse databases directly into the generation process~\cite{10.5555/3495724.3496517}.

Despite these improvements, LLMs remain prone to producing so-called \textit{hallucinations}, i.e., content that is factually or contextually incorrect~\cite{huang2023survey, li-etal-2024-dawn}. Detecting hallucinations is crucial for the safe deployment of LLMs in sensitive fields, as erroneous outputs may seriously erode user trust. 
An effective detector would therefore expand the scope of LLM applications while mitigating risks~\cite{gao2024brief}.

Though many methods address this problem~\cite{sahoo2024comprehensive, 10.1145/3744238}, they often face significant practical constraints, such as the scarcity of annotated datasets~\cite{zhang2023siren} required for supervised methods~\cite{sky2024androids,orgad2025llmsknow}, the high computational cost of generating multiple additional samples~\cite{chen2024inside, hou2025probabilistic}, or the inability of LLMs' output probabilities to fully represent the model's true uncertainty~\cite{fadeeva2024fact,shelmanov-etal-2025-uncertainty}. 


These challenges can be addressed by leveraging LLMs' internal states, which are informative for the hallucination-detection problem~\cite{azaria-mitchell-2023-internal, sriramanan2024llmcheck,gekhman2025insideout}. We introduce TOHA (a TOpology-based HAllucination detector), a training-free method designed for the RAG scenario that leverages the structure of LLM attention maps to identify hallucinations. Our method requires minimal annotated data while avoiding the computational overhead of multiple generations, which makes TOHA both data- and compute-efficient. 

Our core insight is that the prompt-response interconnections within an LLM's attention mechanism reveal the extent to which a response is faithful to the original prompt.
TOHA formalizes this idea by analyzing attention graphs~---~complete weighted graphs derived from LLM attention maps, the representation prior used for topological data analysis (TDA) in NLP~\cite{kushnareva2021artificial, tulchinskii2023topological}. Unlike existing attention-based methods, which treat all attention heads as equally important~\cite{sriramanan2024llmcheck, binkowski2025hallucination} or ignore the geometric structure of attention maps~\cite{sun2025redeep, vazhentsev2025uncertainty}, our approach leverages a small set of attention heads selected by a topological criterion designed to identify hallucinations.

Specifically, TOHA is based on the $\operatorname{MTop-Div}_G(R, P)$, our adaptation of Manifold Topology Divergence ~\cite{barannikov2021manifold} for the graph setting, which quantifies the dissimilarity between the prompt ($P$) and response ($R$) token sets in the attention graph. We demonstrate that this score not only reflects the geometry but also measures the informational novelty of the response relative to the prompt, making it well-suited for hallucination detection in RAG scenarios.

\begin{figure*}[t!]
    \centering
    \includegraphics[width=\linewidth]{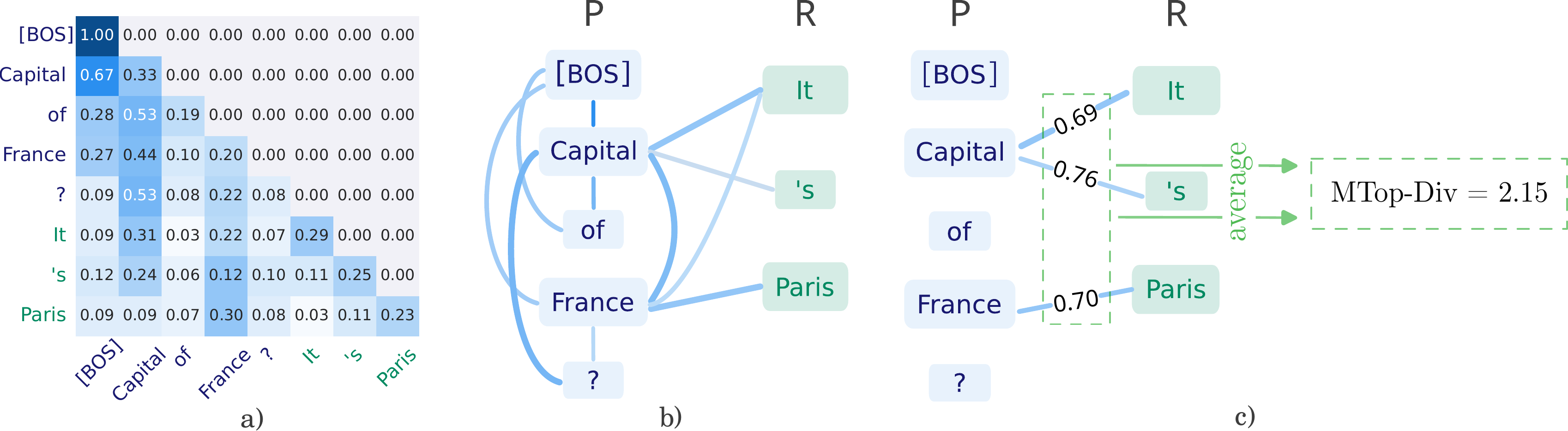} %
    \caption{a) An attention map. Blue and green denote the prompt and response tokens, respectively. b) The corresponding attention graph $G$. Prompt tokens $P$ are located on the left, response tokens $R$~---~on the right. To keep the figure neat, we only plot edges with an attention score of at least $0.15$. c) The minimum spanning forest attaching $R$ to $P$ and the corresponding $\operatorname{MTop-Div}$ value.} \label{fig:attention_graph}
\end{figure*}

By analyzing divergence values across attention heads, we identified a subset of heads that consistently yield higher scores for hallucinated samples (Figure~\ref{fig:average_distances}), irrespective of the dataset. 
TOHA's final hallucination score is the average $\operatorname{MTop-Div}_G(R, P)$ value from these ``hallucination-aware'' heads. Some of these heads are associated with copying behavior, which is aligned with prior findings~\cite{sun2025redeep}. 

Here are our main \textbf{contributions}:
\begin{itemize}
    \item We introduce TOHA, a training-free framework for hallucination detection in the RAG scenario that leverages the topology of LLM attention graphs. Our method operates up to an order of magnitude faster than methods of comparable quality and requires minimal annotated data. 
    \item Central to TOHA is $\operatorname{MTop-Div}_G(R, P)$, an adaptation of Manifold Topology Divergence for graphs, which quantifies the topological dissimilarity between prompt and response sets of tokens. We demonstrate that this score characterizes the novelty of the response relative to the prompt, making it highly effective for detecting hallucinations in RAG systems.
    \item By analyzing the proposed score, we discover the existence of ``hallucination-aware'' attention heads, which consistently yield greater divergence values for hallucinated samples across different datasets. This finding ensures TOHA's efficiency and strong cross-domain transferability: averaging $\operatorname{MTop-Div}_G(R, P)$ over just 10 heads is sufficient for robust hallucination detection. 
    \item Our experiments show that TOHA consistently matches or exceeds state-of-the-art performance across multiple benchmarks when applied to open-source LLMs of varying sizes, including 7B- and 13B-parameter models.
    
\end{itemize}

\section{Background}

\subsection{Attention Map as a Weighted Graph}
Modern LLMs are mainly based on the self-attention mechanism~\cite{vaswani2017attention}. 
Let $X \in \mathbb{R}^{n \times d}$ be a matrix consisting of $d$-dimensional representations of $n$ tokens, $W_Q, \, W_K, \, W_V \in \mathbb{R}^{d \times d}$ be trainable projection matrices. Given a set of queries $Q = XW_Q \in \mathbb{R}^{n \times d}$, a set of keys $K=XW_K  \in \mathbb{R}^{n \times d}$, and corresponding values $V = X W_V  \in \mathbb{R}^{n \times d}$, the attention mechanism calculates a weighted sum of the values:
\begin{equation}
    \operatorname{Attention}(Q, K, V) = W(Q, K) V,
\end{equation}
where $W(Q, K)$ is an attention map 
\begin{equation}
    \operatorname{W}(Q, K) = \operatorname{softmax} \left(\frac{Q K^T}{\sqrt{d}}\right),
\end{equation}
and each entry $w_{ij} = W_{ij}(Q, K)$ captures how strongly token $i$ attends to token $j$, with greater values indicating closer relationship. We consider decoder-only LLMs, for which the attention maps are lower triangular.

An attention map can be reframed as a complete undirected weighted graph $G$ with edge weights $1-w_{ij}$ to represent pseudo-distances between tokens. We call $G$ an $\textit{attention graph}$. It naturally partitions into prompt ($P$) and response ($R$) tokens (see Figure~\ref{fig:attention_graph}b). We analyze the topological relationships between these node subsets, which, as we assume, should be indicative of hallucinations in the RAG scenario.



\subsection{Manifold Topology Divergence}

The Manifold Topology Divergence, or $\operatorname{MTop-Div}$, was proposed in~\cite{barannikov2021manifold}.
$\operatorname{MTop-Div}(M, N)$ quantifies the difference between two data manifolds $\mathcal{M}$ and $\mathcal{N}$, approximated by point clouds $M$ and $N$. It is computed as the sum of interval lengths in the $\operatorname{Cross-Barcode}(M, N)$, a set of intervals representing topological features distinguishing $N$ from $M \cup N$. A larger divergence indicates a greater topological difference between the manifolds. Appendix~\ref{sec:appendix_topology} contains more details on TDA.

\section{Method}\label{sec:method}

This section introduces the $\operatorname{MTop-Div}_G(R, P)$ score, which is designed to quantify the topological divergence between the prompt and response subgraphs in attention maps. We demonstrate that not only can it be interpreted as a geometric characteristic of the attention graph, but also as an information-theoretic measure of the novelty of the response in relation to the prompt.  With $\operatorname{MTop-Div}_G(R, P)$, we identify ``hallucination-aware'' heads that consistently separate hallucinated samples from grounded, irrespective of the dataset. This finding allows us to formulate the TOHA algorithm (Algorithm~\ref{alg:toha}), which computes a final hallucination score by averaging the divergence values from these specific heads. 

\subsection{$\operatorname{MTop-Div}$ for Attention Graphs: Definition}

Let $R$ and $P$ be the response and prompt vertex sets in an attention graph $G$. After zeroing edge weights between the $P$ nodes, we compute the $0$-th order homology barcode $\mathcal{B}_0$ of the Vietoris-Rips simplicial complex of the modified graph. Essentially, this barcode tracks the evolution of connected components as we threshold the graph edges (see Appendix~\ref{sec:appendix_topology} for details).

Our proposed topological divergence is a sum of the lengths of $\mathcal{B}_0$ intervals:
\begin{equation}
     \operatorname{MTop-Div}_G(R, P) = \sum\limits_{\scriptstyle [b_i, d_i] \in \mathcal{B}_0} |d_i - b_i|. \nonumber
\end{equation}

This score can be interpreted from two perspectives: geometric (as the length of a minimal spanning forest in the attention graph) and information-theoretic (as a measure of the response novelty in the space induced by query-key matrices).  We explore these two interpretations in the following subsections.

\subsection{$\operatorname{MTop-Div}_G(R, P)$ and the Geometry of the Attention Graph}~\label{sec:mtd_geometry}

Formally, we can prove the following property (see proof in Appendix~\ref{appendProof31}):

\begin{proposition}\label{prop:stability_properties}
Consider an attention graph $G$ with vertex set $V_G$ and its complementary vertex subsets $P, R$,  where $P \cup R = V_G$ and $P \cap R = \varnothing$. $\operatorname{MTop-Div}(R, P)$  value equals the length of the minimal spanning forest (MSF) attaching $R$ to $P$. 
\end{proposition}

This proposition's geometric meaning is illustrated in Figure~\ref{fig:attention_graph}. While the connections between prompt nodes (Figure~\ref{fig:attention_graph}b) are semantically and syntactically meaningful, we hypothesize they primarily introduce noise for hallucination detection (see Section~\ref{sec:ablation} for the corresponding experiment). After setting these distances to zero, we construct an MSF on the modified graph (Figure~\ref{fig:attention_graph}c). The sum of the lengths of the remaining edges in this MSF is precisely the value of $\operatorname{MTop-Div}_G(R, P)$. 




\subsection{$\operatorname{MTop-Div}_G(R, P)$ as a Topology-Based Novelty Score}
Consider an attention graph $G$. 
It can be interpreted as a non-metric space, with ``distances'' induced by the attention weights.
By Proposition 3.1, $\operatorname{MTop-Div}_G(R, P)$ equals the length of the minimum spanning forest attaching the response tokens $R$ to the prompt tokens $P$. Therefore,
\[
\operatorname{MTop-Div}_G(R, P)\ge L_{\mathrm{MST}}(R\cup P)-L_{\mathrm{MST}}(P),
\]
since adding an MST on $P$ to such a forest yields a spanning tree on $R\cup P$.

This shows that $\mathrm{MTop\text{-}Div}_G(R, P)$ is at least the increase in MST length obtained when the response tokens are added to the prompt tokens. Recall that MST length is commonly used as a proxy for geometric dispersion~\cite{Mller2012}, more precisely, the entropy estimate given the MST of length $L$ on points $\mathcal{X}$ is the following:
$$H_{MST}(\mathcal{X}) = d\log L - (d - 1)\log n + \log \beta_d,$$
where $|\mathcal{X}| = n$, $d$ is the intrinsic dimensionality of the data, and $\beta_d$ is some data-independent constant. 
Therefore larger values of $\mathrm{MTop\text{-}Div}_G(R, P)$ can be interpreted as indicating greater structural novelty of the response relative to the prompt. Thus, our $\operatorname{MTop-Div}_G(R,  P)$ should be an effective statistic for identifying hallucinated responses.

\subsection{Hallucination-Aware Heads}


Denote by $h_{ij}$ the $j$-th attention head from the layer $i$. For the specific data sample $s$ and head $h_{ij}$, let $G^s_{ij}$ be the corresponding attention graph, $P^s_{ij}, R^s_{ij}$~---~its prompt and response vertex subsets. 

\begin{figure}[h]
    \centering
    \includegraphics[width=\columnwidth]{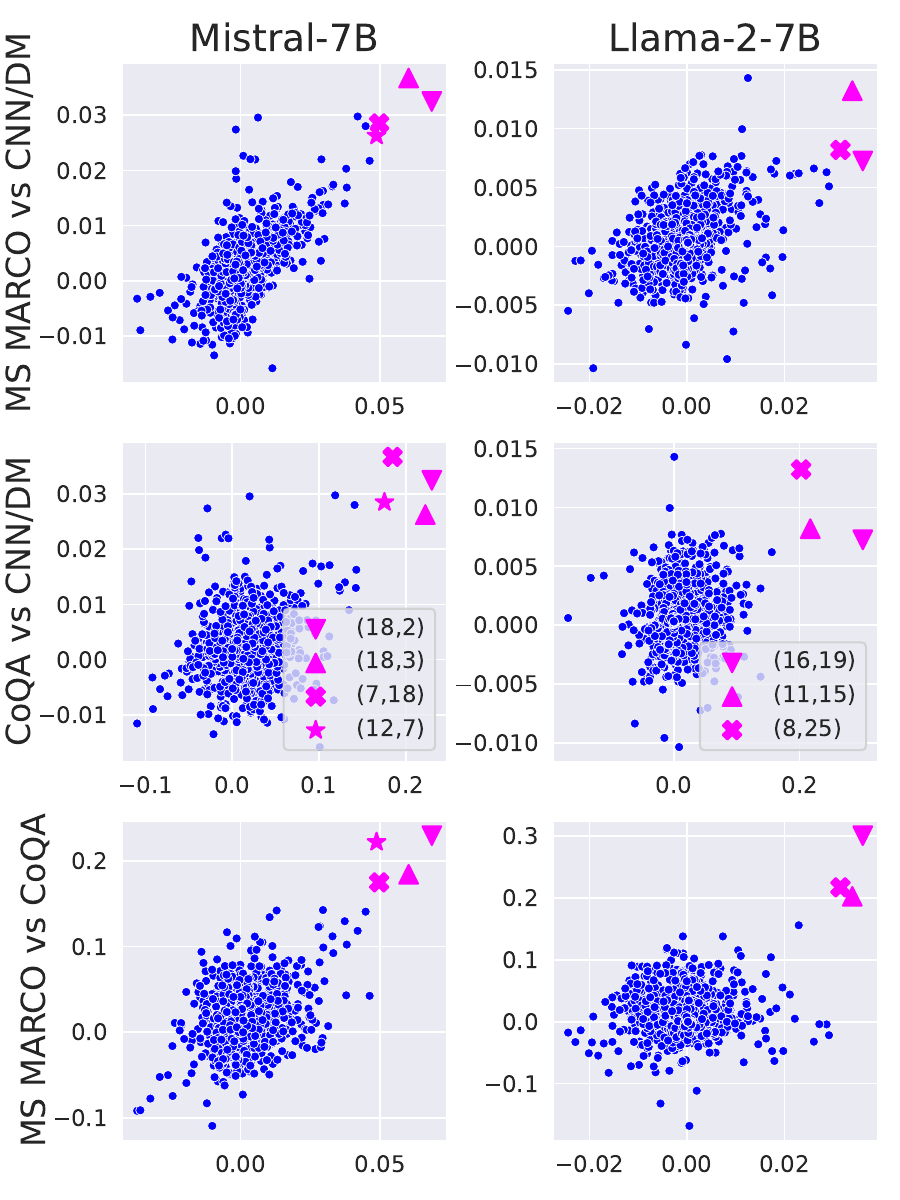} %
    \caption{$\Delta_{ij}$ values for $ij$-th heads. Vertical axis corresponds to the difference on dataset (B), horizontal~---~to the one on dataset (A). The heads that separate samples best are highlighted in pink; their (layer, head) positions are reflected in the legend. } \label{fig:average_distances}
\end{figure}

We examined typical values of the average distance between hallucinated and grounded training examples for different heads and layers:
\begin{equation}\label{eq:delta_ij}
    \Delta_{ij} = \frac{1}{|S_{\mathrm{hallu}}|}\sum\limits_{s 
\in S_{\mathrm{hallu}}}d_{ij}(s) - \frac{1}{|S_{\mathrm{gr}}|}\sum\limits_{s 
\in S_{\mathrm{gr}}}d_{ij}(s), 
\end{equation}
where $S_{\mathrm{hallu}}$ stands for all hallucinated samples from the training set, $S_{\mathrm{gr}}$ stands for all grounded training samples, and
$$
d_{ij}(s) = \frac{1}{|R_{ij}^s|}\operatorname{MTop-Div}_{G^s_{ij}}(R^s_{ij}, P^s_{ij}).
$$
Figure~\ref{fig:average_distances} displays sample differences $\Delta_{ij}$ across three datasets, with each marker representing some attention head. We discovered that the same four (for Mistral-7B) and three (for Llama-2-7B) heads,  highlighted in pink, demonstrate similar behavior across the datasets:  they consistently appear in the upper-right corner, indicating strong separation between hallucinated and grounded samples, irrespective of the dataset. This finding suggests that there are specific attention heads somehow ``aware'' of the presence of hallucination, which is captured by the proposed $\operatorname{MTop-Div}_G(R, P)$.
\subsection{TOHA}
The existence of ``hallucination-aware'' attention heads underlies our method, which is detailed in Algorithm~\ref{alg:toha}. TOHA uses two annotated probe sets ($S_h$ for hallucinations, $S_g$ for grounded samples) to rank attention heads by their separation capability $\Delta_{ij}$ and select the top $N_{\mathrm{opt}}$ ones with the combined probe size $|S_h \cup S_g|$  kept small (see Figure~\ref{fig:probe_set_size_ablation}). During testing, the final hallucination score for test samples $T$ equals the average topological divergence from the selected $N_{\mathrm{opt}}$ heads. For computational efficiency, we limit \( N_{\mathrm{opt}} \) to a maximum of $N_{\max} = 10$ in all experiments.

\begin{algorithm}[t]
\caption{TOHA Algorithm}\label{alg:toha}
\begin{algorithmic}[1]
\Require{$d_{ij}(s)$, $S_h$, $S_g$, $V = S_h \cup S_g$, $T$, $N_{\max}$}
\Ensure{Hallucination scores $\{p_s\}$ for $s \in T$}
\Statex
\Function{HeadsSelection}{}
    \For{each head $h_{ij}$}
        \State Calculate $\Delta_{ij}$ according to eq.~\eqref{eq:delta_ij}
    \EndFor
    \State $H \gets$ sort all heads by $\Delta_{ij}$ (descending)
    \State $N_{\text{opt}}, \text{AUROC}_{\max} \gets 1, 0$
    \State Initialize $p_s \gets 0$ for all $s \in V$
    \For{$N = 1$ \textbf{to} $N_{\max}$}
        \For{each $s \in V$}
            \State $p_s \gets \frac{N-1}{N}p_s + \frac{1}{N}d_{h_N}(s)$
        \EndFor
        \State $\text{auroc} \gets \text{AUROC}(\{y_s\}_{s \in V}, \{p_s\}_{s \in V})$
        \If{$\text{auroc} > \text{AUROC}_{\max}$}
            \State $\text{AUROC}_{\max} \gets \text{auroc}; \ N_{\text{opt}} \gets N$
        \EndIf
    \EndFor
    \State \Return $N_{\text{opt}}$
\EndFunction
\Statex
\Function{Prediction}{$N_{\text{opt}}$}
    \For{each $s \in T$}
        \State $p_s \gets \frac{1}{N_{\text{opt}}} \sum_{i=1}^{N_{\text{opt}}} d_{h_i}(s)$
    \EndFor
    \State \Return $\{p_s\}$
\EndFunction
\end{algorithmic}
\end{algorithm}
\section{Results}\label{sec:experiments}
\subsection{Experiment Setting}
\paragraph{Datasets and models.} We consider five tasks that evaluate question answering and summarization abilities of LLMs: two benchmarks of RAGTruth~\cite{niu2023ragtruth}~---~long-form QA dataset MS MARCO and summarization dataset CNN/DM; conversational QA dataset CoQA~\cite{reddy2019coqa}, reading comprehension dataset SQuAD~\cite{rajpurkar2016squad}, and extreme summarization dataset XSum~\cite{narayan-etal-2018-dont}. For more details, see Appendix~\ref{sec:appendix_datasets}.

We employ five popularly adopted open-source LLMs with accessible inner states: LLaMA-2-7B-chat, LLaMA-2-13B-chat, LLaMA-3.1-8B-Instruct, Mistral-7B-Instruct-v0.1, and Qwen2.5-7B-Instruct. 
As the RAGTruth dataset does not contain responses for LLaMA-3.1-8B and Qwen-2.5-7B, we conducted experiments on SQuAD, CoQA, and XSum for these models. 

\paragraph{Baselines.} We compare TOHA with a comprehensive set of eight baselines: uncertainty-based perplexity~\cite{ren2023outofdistribution} and max entropy~\cite{fadeeva2024fact}; inner states-based ReDeEP~\cite{sun2025redeep}, HaloScope~\cite{du2024haloscope}, and LLM-Check~\cite{sriramanan2024llmcheck}; consistency-based semantic entropy~\cite{farquhar2024detecting}, EigenScore~\cite{chen2024inside}, and SelfCheckGPT~\cite{manakul2023selfcheckgpt}.

\paragraph{Implementation details. } The reported results are averaged over 5 runs with different data splits, using test sets comprising 25\% of the data and a fixed validation set size of 100, following the methodology of HaloScope~\cite{du2024haloscope}. To ensure a fair comparison, we consider two implementations of the latter: a standard setting with $20$  generations (see Tables~\ref{tab:additional_generations}-\ref{tab:additional_generations_newer_models}), and an efficiency-comparable one with minimal number of generations~---~$1$ for SelfCheckGPT, $5$ for semantic entropy and EigenScore (Tables~\ref{tab:mistral}-\ref{tab:llama_3_8b}). Appendix~\ref{sec:appendix_implementation} provides additional implementation details. 

\defcitealias{manakul2023selfcheckgpt}{[1]}
\defcitealias{farquhar2024detecting}{[2]}
\defcitealias{chen2024inside}{[3]}
\defcitealias{du2024haloscope}{[4]}
\defcitealias{sriramanan2024llmcheck}{[5]}
\defcitealias{ren2023outofdistribution}{[6]}
\defcitealias{fadeeva2024fact}{[7]}
\defcitealias{sun2025redeep}{[8]}

\begin{table*}[t!]
\centering
\caption{ROC AUC $(\uparrow)$ of hallucination detection techniques for three LLMs. The best results for each model are highlighted in \textbf{bold}, and the second best are \underline{underlined}. }
 \label{tab:mistral}
 \renewcommand{\arraystretch}{1} 
\resizebox{\textwidth}{!}{%
\begin{tabular}{lccccccc}
\hline
\multirow{2}{*}{Method} & Single &\multirow{2}{*}{MS MARCO} & CNN/DM + &  \multirow{2}{*}{CoQA} &  \multirow{2}{*}{SQuAD} &  \multirow{2}{*} {XSum} \\ 
 & generation & & Recent News &  & & & \\ \hline
\multicolumn{7}{c}{Mistral-7B} \\ \hline
SelfCheckGPT~\citetalias{manakul2023selfcheckgpt} & \xmark & 0.63 \footnotesize{$\pm$ 0.04} & 0.51 \footnotesize{$\pm$ 0.04} & \underline{0.86 \footnotesize{$\pm$ 0.02}} & 0.71 \footnotesize{$\pm$ 0.04} & \underline{0.66 \footnotesize{$\pm$ 0.04}} \\
Semantic entropy~\citetalias{farquhar2024detecting} & \xmark & 0.54 \footnotesize{$\pm$ 0.03} & 0.51 \footnotesize{$\pm$ 0.04} & 0.83 \footnotesize{$\pm$ 0.02} & 0.70 \footnotesize{$\pm$ 0.03} & 0.56 \footnotesize{$\pm$ 0.03} \\
EigenScore~\citetalias{chen2024inside} & \xmark & 0.54 \footnotesize{$\pm$ 0.04} & 0.50 \footnotesize{$\pm$ 0.06} & 0.74 \footnotesize{$\pm$ 0.02} & 0.71 \footnotesize{$\pm$ 0.04} & 0.58 \footnotesize{$\pm$ 0.04} \\
HaloScope~\citetalias{du2024haloscope} & \cmark & 0.57 \footnotesize{$\pm$ 0.08} & 0.51 \footnotesize{$\pm$ 0.10} & 0.62 \footnotesize{$\pm$ 0.08} & \underline{0.92 \footnotesize{$\pm$ 0.07}} & 0.62 \footnotesize{$\pm$ 0.02} \\
LLM-Check~\citetalias{sriramanan2024llmcheck} & \cmark & 0.49 \footnotesize{$\pm$ 0.03} & 0.49 \footnotesize{$\pm$ 0.03} & 0.60 \footnotesize{$\pm$ 0.01} & 0.50 \footnotesize{$\pm$ 0.03} & 0.58 \footnotesize{$\pm$ 0.04} \\
Perplexity~\citetalias{ren2023outofdistribution} & \cmark & 0.45 \footnotesize{$\pm$ 0.01} & \underline{0.54 \footnotesize{$\pm$ 0.02}} & 0.54 \footnotesize{$\pm$ 0.03} & 0.81 \footnotesize{$\pm$ 0.05} & 0.54 \footnotesize{$\pm$ 0.06} \\
Max entropy~\citetalias{fadeeva2024fact} & \cmark & \underline{0.68 \footnotesize{$\pm$ 0.04}} & \textbf{0.60 \footnotesize{$\pm$ 0.07}} & 0.73 \footnotesize{$\pm$ 0.00} & 0.75 \footnotesize{$\pm$ 0.05} & \textbf{0.71 \footnotesize{$\pm$ 0.02}} \\
ReDEEP~\citetalias{sun2025redeep} & \cmark & 0.54 \footnotesize{$\pm$ 0.02} & 0.47 \footnotesize{$\pm$ 0.06} & 0.59 \footnotesize{$\pm$ 0.03} & 0.45 \footnotesize{$\pm$ 0.05} & 0.63 \footnotesize{$\pm$ 0.04} \\
TOHA (ours) & \cmark & \textbf{0.76 \footnotesize{$\pm$ 0.04}} & 
\textbf{0.60 \footnotesize{$\pm$ 0.09}} & \textbf{0.89 \footnotesize{$\pm$ 0.01}} & \textbf{0.96 \footnotesize{$\pm$ 0.01}} & \underline{0.66 $\pm$ 0.05} \\ \hline
\multicolumn{7}{c}{LLama-2-7B} \\ \hline
SelfCheckGPT~\citetalias{manakul2023selfcheckgpt} & \xmark & \underline{0.59 \footnotesize{$\pm$ 0.03}} & \textbf{0.60 \footnotesize{$\pm$ 0.03}} & 0.66 \footnotesize{$\pm$ 0.03} & 0.57 \footnotesize{$\pm$ 0.03} & \underline{0.64 \footnotesize{$\pm$ 0.05}} \\
Semantic entropy~\citetalias{farquhar2024detecting} & \xmark & 0.53 \footnotesize{$\pm$ 0.03} & 0.51 \footnotesize{$\pm$ 0.03} & \underline{0.76 \footnotesize{$\pm$ 0.01}} & 0.73 \footnotesize{$\pm$ 0.03} & 0.61 \footnotesize{$\pm$ 0.04} \\
EigenScore~\citetalias{chen2024inside} & \xmark & 0.55 \footnotesize{$\pm$ 0.03}  & 0.53 \footnotesize{$\pm$ 0.04} & 0.61 \footnotesize{$\pm$ 0.03} & \underline{0.75 \footnotesize{$\pm$ 0.02}} & 0.63 \footnotesize{$\pm$ 0.02} \\
HaloScope~\citetalias{du2024haloscope} & \cmark & 0.51 \footnotesize{$\pm$ 0.05} & 0.48 \footnotesize{$\pm$ 0.05} & 0.61 \footnotesize{$\pm$ 0.04} & 0.67 \footnotesize{$\pm$ 0.04} & 0.57 \footnotesize{$\pm$ 0.07} \\
LLM-Check~\citetalias{sriramanan2024llmcheck} & \cmark & 0.44 \footnotesize{$\pm$ 0.02} & 0.49 \footnotesize{$\pm$ 0.06} & 0.60 \footnotesize{$\pm$ 0.03} & 0.49 \footnotesize{$\pm$ 0.01} & 0.61 \footnotesize{$\pm$ 0.01} \\
Perplexity~\citetalias{ren2023outofdistribution} & \cmark & 0.54 \footnotesize{$\pm$ 0.04} & 0.44 \footnotesize{$\pm$ 0.03} & 
0.74 \footnotesize{$\pm$ 0.02} & 0.46 \footnotesize{$\pm$ 0.03} & 0.56 \footnotesize{$\pm$ 0.09} \\
Max entropy~\citetalias{fadeeva2024fact} & \cmark & 
\textbf{0.65 \footnotesize{$\pm$ 0.04}} & \underline{0.59 \footnotesize{$\pm$ 0.06}} & 0.65 \footnotesize{$\pm$ 0.03} & 0.73 \footnotesize{$\pm$ 0.04} & 0.56 \footnotesize{$\pm$ 0.03} \\
ReDEEP~\citetalias{sun2025redeep} & \cmark & 0.54 \footnotesize{$\pm$ 0.04} & 0.52 \footnotesize{$\pm$ 0.04}  & 0.72 \footnotesize{$\pm$ 0.04} & 0.42 \footnotesize{$\pm$ 0.08} & 0.54 \footnotesize{$\pm$ 0.06}  \\
TOHA (ours) & \cmark & \textbf{0.65 \footnotesize{$\pm$ 0.02}} & 0.56 \footnotesize{$\pm$ 0.02} & \textbf{0.90 \footnotesize{$\pm$ 0.01}} & \textbf{0.87 \footnotesize{$\pm$ 0.04}} & \textbf{0.68 \footnotesize{$\pm$ 0.05}} \\ \hline
\multicolumn{7}{c}{LLaMA-2-13B} \\ \hline
SelfCheckGPT~\citetalias{manakul2023selfcheckgpt} & \xmark & 0.58 \footnotesize{$\pm$ 0.04} & \textbf{0.58 \footnotesize{$\pm$ 0.05}} & \underline{0.77 \footnotesize{$\pm$ 0.02}} & 0.64 \footnotesize{$\pm$ 0.03} & \underline{0.60 \footnotesize{$\pm$ 0.04}} \\
Semantic entropy~\citetalias{farquhar2024detecting} & \xmark & 0.57 \footnotesize{$\pm$ 0.04} & 0.54 \footnotesize{$\pm$ 0.03} & 0.76 \footnotesize{$\pm$ 0.04} & 0.65 \footnotesize{$\pm$ 0.03} & \underline{0.60 \footnotesize{$\pm$ 0.03}} \\
EigenScore~\citetalias{chen2024inside} & \xmark & 0.56 \footnotesize{$\pm$ 0.04} & 0.47 \footnotesize{$\pm$ 0.04} & 0.57 \footnotesize{$\pm$ 0.03} & 0.57 \footnotesize{$\pm$ 0.02} & 0.52 \footnotesize{$\pm$ 0.06} \\
HaloScope~\citetalias{du2024haloscope} & \cmark & 0.54 \footnotesize{$\pm$ 0.09} & 0.51 \footnotesize{$\pm$ 0.04} & 0.57 \footnotesize{$\pm$ 0.03} & 0.55 \footnotesize{$\pm$ 0.02} & 0.55 \footnotesize{$\pm$ 0.07} \\
LLM-Check~\citetalias{sriramanan2024llmcheck} & \cmark & 0.49 \footnotesize{$\pm$ 0.06} & \underline{0.56 \footnotesize{$\pm$ 0.05}} & 0.57 \footnotesize{$\pm$ 0.02} & 0.57 \footnotesize{$\pm$ 0.07} & 0.57 \footnotesize{$\pm$ 0.07} \\
Perplexity~\citetalias{ren2023outofdistribution} & \cmark & 0.54 \footnotesize{$\pm$ 0.04} & 0.46 \footnotesize{$\pm$ 0.07} & 0.62 \footnotesize{$\pm$ 0.03} & 0.45 \footnotesize{$\pm$ 0.02} & 0.49 \footnotesize{$\pm$ 0.05} \\
Max entropy~\citetalias{fadeeva2024fact} & \cmark & \underline{0.62 \footnotesize{$\pm$ 0.03}} & 0.53 \footnotesize{$\pm$ 0.06} & 0.66 \footnotesize{$\pm$ 0.03} & \underline{0.78 \footnotesize{$\pm$ 0.02}} & 0.59 \footnotesize{$\pm$ 0.04} \\
ReDEEP~\citetalias{sun2025redeep} & \cmark & \underline{0.62 \footnotesize{$\pm$ 0.06}} & 0.48 \footnotesize{$\pm$ 0.05} & 0.73 \footnotesize{$\pm$ 0.02} & 0.48 \footnotesize{$\pm$ 0.07}  & 0.58 \footnotesize{$\pm$ 0.08} \\
TOHA (ours) & \cmark & \textbf{0.67 \footnotesize{$\pm$ 0.04}} & \underline{0.56 \footnotesize{$\pm$ 0.05}} & \textbf{0.92 \footnotesize{$\pm$ 0.02}} & \textbf{0.88 \footnotesize{$\pm$ 0.05}} & \textbf{0.66 \footnotesize{$\pm$ 0.03}} \\
 \hline
\end{tabular}}
\end{table*}

\subsection{Results}

\paragraph{Main results.} The results of our experiments are provided in Tables~\ref{tab:mistral}--\ref{tab:llama_3_8b}.
We evaluate TOHA against state-of-the-art hallucination detection methods and demonstrate its competitive performance
TOHA significantly outperforms uncertainty-based baselines and matches the quality of consistency-based approaches, achieving notable improvements of $11.7\%$  on the challenging long-form QA MS MARCO dataset for Mistral-7B and $21.6\%$ on the conversational QA dataset CoQA for LLama-2-7B.

We rigorously validate the statistical significance of our results using a critical difference diagram with Wilcoxon-Holm post-hoc analysis~\cite{IsmailFawaz2018deep}. The results provided in Tables~\ref{tab:pairwise_pvalues}-\ref{tab:ranking} (Appendix~\ref{sec:appendix_other_results}) confirm that while many baseline methods remain statistically indistinguishable from one another, TOHA achieves the top overall rank (1.67) and its performance improvements are statistically significant compared to every other evaluated method ($p \le 0.0016$).

\begin{figure*} [h!] \hspace{-2mm}\includegraphics[width=1.03\textwidth]{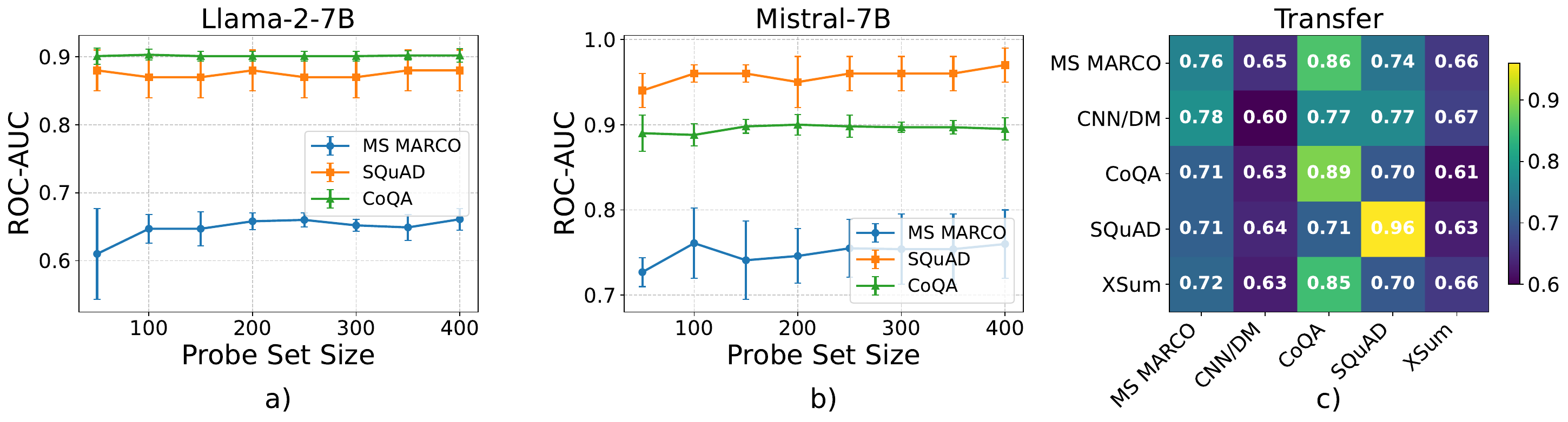} %
    \caption{(a)-(b): Detection quality dependence on the size of a probe set, models: Mistral-7B (left), LLama-2-7B (right). (c) Generalizability between the datasets, model: Mistral-7B. The vertical axis corresponds to the origin of the probe set, and the horizontal axis to the test dataset. } \label{fig:probe_set_size_ablation}
\end{figure*}

To evaluate TOHA's robustness to data distribution change, we conducted transfer experiments on Mistral-7B, see Figure~\ref{fig:probe_set_size_ablation}(c) for the results. TOHA exhibits strong transferability: across the XSum and CNN/DM datasets, performance falls within the method's standard deviation (see the statistical significance analysis in Table~\ref{tab:pvalues}), while remaining competitive on the other datasets (Table~\ref{tab:mistral}).

\paragraph{Evaluation on a multi-hop dataset.} To validate TOHA in a more realistic setting, we consider an additional experiment using the HotpotQA~\cite{yang2018hotpotqa} dataset.
It consists of questions that require knowledge from multiple supporting documents~---~much like real-world queries, which rarely rely on a single source of information.
The results are provided in Table~\ref{tab:hotpot_qa}. 
TOHA demonstrates superior performance to baselines, confirming its effectiveness ``in the wild''.

\begin{figure*}[t]
    \centering
    \includegraphics[width=\textwidth]{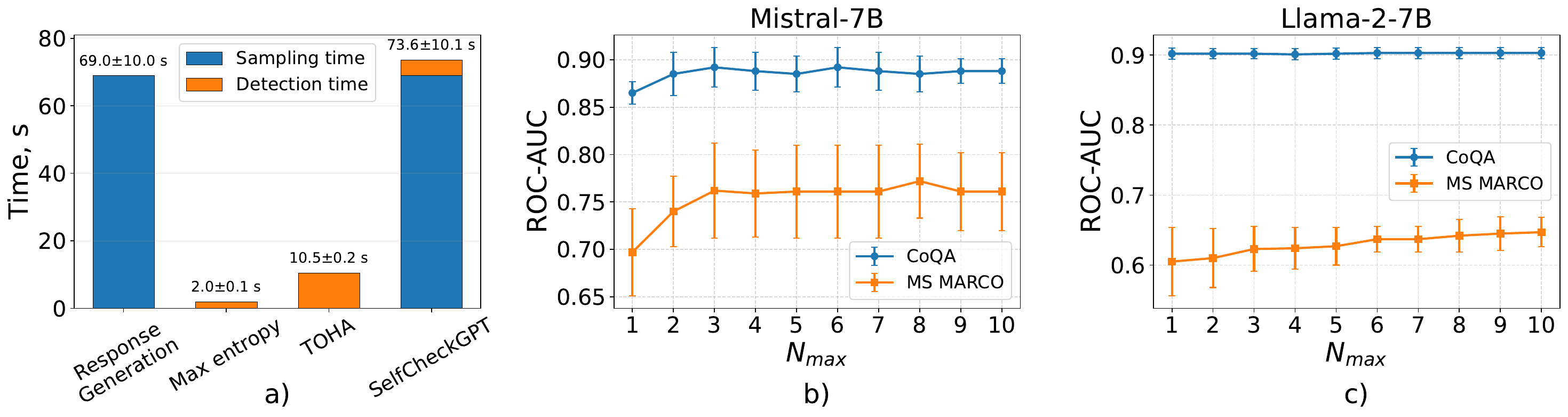} 
    \caption{a): Inference time comparison (seconds) for various methods evaluated on 16 MS MARCO samples using Mistral-7B. SelfCheckGPT measurement includes one additional generated answer per sample. b)-c): ROC-AUC performance of TOHA across different numbers of selected attention heads ($N_{max}$) on Mistral-7B and Llama-2-7B. } \label{fig:n_max_comparison}
\end{figure*}

\paragraph{Efficiency.}  Figure~\ref{fig:n_max_comparison}(a) shows that TOHA is approximately seven times faster than SelfCheckGPT with a \textit{single} additional generation. Given that SelfCheckGPT typically requires 10-20 generations, this makes TOHA over 70 times faster in practice. Furthermore, TOHA's runtime is close to the lightweight entropy baseline; with a more efficient low-level implementation, it has the potential to approach the cost of a single forward pass while achieving higher accuracy than other inexpensive techniques.

\begin{table}[t!]
\centering
\caption{ROC AUC $(\uparrow)$ of hallucination detection techniques. The best results for each model are highlighted in \textbf{bold}, and the second best are \underline{underlined}.}
 \label{tab:llama_3_8b}
\resizebox{\columnwidth}{!}{\begin{tabular}{lcccc}
\hline
Method & Single gen. & CoQA & SQuAD & XSum \\ \hline
\multicolumn{5}{c}{LLaMA-3.1-8B} \\ \hline
SelfCheckGPT~\citetalias{manakul2023selfcheckgpt} & \xmark & \textbf{0.91 \footnotesize{$\pm$ 0.01}} & 0.65 \footnotesize{$\pm$ 0.05} & \textbf{0.68 \footnotesize{$\pm$ 0.05}} \\
Semantic entropy~\citetalias{farquhar2024detecting} & \xmark & 0.78 \footnotesize{$\pm$ 0.02} & 0.55 \footnotesize{$\pm$ 0.07} & 0.47 \footnotesize{$\pm$ 0.04} \\
EigenScore~\citetalias{chen2024inside} & \xmark & 0.80 \footnotesize{$\pm$ 0.02} & 0.45 \footnotesize{$\pm$ 0.05} & 0.50 \footnotesize{$\pm$ 0.03} \\
HaloScope~\citetalias{du2024haloscope} & \cmark & 0.67 \footnotesize{$\pm$ 0.08} & \underline{0.84 \footnotesize{$\pm$ 0.07}}  & 0.55 \footnotesize{$\pm$ 0.06}  \\
LLM-Check~\citetalias{sriramanan2024llmcheck} & \cmark & 0.54 \footnotesize{$\pm$ 0.06}  & 0.49 \footnotesize{$\pm$ 0.05} & 0.52 \footnotesize{$\pm$ 0.04} \\
Perplexity~\citetalias{ren2023outofdistribution} & \cmark & 0.51 \footnotesize{$\pm$ 0.05} & 0.68 \footnotesize{$\pm$ 0.02} & \underline{0.65 \footnotesize{$\pm$ 0.03}} \\
Max entropy~\citetalias{fadeeva2024fact} & \cmark  & 0.82  \footnotesize{$\pm$ 0.03} & 0.60 \footnotesize{$\pm$ 0.02} & \underline{0.65 \footnotesize{$\pm$ 0.03}}  \\
ReDEEP~\citetalias{sun2025redeep} & \cmark & 0.58 \footnotesize{$\pm$ 0.08} & 0.39 \footnotesize{$\pm$ 0.04} & 0.62 \footnotesize{$\pm$ 0.06}\\ 
TOHA (ours) & \cmark & \underline{0.84 \footnotesize{$\pm$ 0.01}} & \textbf{0.87 \footnotesize{$\pm$ 0.03}} & \underline{0.65 \footnotesize{$\pm$ 0.05}} \\ \hline
\multicolumn{5}{c}{Qwen2.5-7B} \\ \hline
SelfCheckGPT~\citetalias{manakul2023selfcheckgpt} & \xmark & 0.69 \footnotesize{$\pm$ 0.01}  & 0.74 \footnotesize{$\pm$ 0.02} & \textbf{0.69 \footnotesize{$\pm$ 0.02}}  \\
Semantic entropy~\citetalias{farquhar2024detecting} & \xmark & 0.76 \footnotesize{$\pm$ 0.04} &  0.57 \footnotesize{$\pm$ 0.04} & 0.64 \footnotesize{$\pm$ 0.03}\\
EigenScore~\citetalias{chen2024inside} & \xmark & 0.78 \footnotesize{$\pm$ 0.03} & 0.55 \footnotesize{$\pm$ 0.04}  & 0.55 \footnotesize{$\pm$ 0.04}\\
HaloScope~\citetalias{du2024haloscope} & \cmark& 0.66 \footnotesize{$\pm$ 0.13} & \underline{0.75 \footnotesize{$\pm$ 0.04}}  & 0.57 \footnotesize{$\pm$ 0.07} \\
LLM-Check~\citetalias{sriramanan2024llmcheck} & \cmark & 0.53 \footnotesize{$\pm$ 0.07} &  0.54 \footnotesize{$\pm$ 0.06} & 0.54 \footnotesize{$\pm$ 0.02} \\
Perplexity~\citetalias{ren2023outofdistribution} & \cmark & 0.39 \footnotesize{$\pm$ 0.03} & 0.65 \footnotesize{$\pm$ 0.03}  & 0.53 \footnotesize{$\pm$ 0.05}\\
Max entropy~\citetalias{fadeeva2024fact} & \cmark & \textbf{0.85 \footnotesize{$\pm$ 0.02}} & 0.47 \footnotesize{$\pm$ 0.05}& 0.60 \footnotesize{$\pm$ 0.06} \\
ReDEEP~\citetalias{sun2025redeep} & \cmark & 0.37 \footnotesize{$\pm$ 0.10} & 0.56 \footnotesize{$\pm$ 0.03} & \underline{0.67 \footnotesize{$\pm$ 0.03}}\\ 
TOHA (ours)&  \cmark & \underline{0.79 \footnotesize{$\pm$ 0.05}} & \textbf{0.77 \footnotesize{$\pm$ 0.02}}  & \textbf{0.69 \footnotesize{$\pm$ 0.03}}\\ \hline
\end{tabular}}
\end{table}

\begin{table}
\caption{ROC-AUC $(\uparrow)$  values on the HotpotQA dataset. Best results are highlighted in \textbf{bold}, and the second best are \underline{underlined}.}
\resizebox{\columnwidth}{!}{\begin{tabular}{lccc}
\hline
Method & Single gen. & Mistral-7B & Llama-2-13B \\
\hline
SelfCheckGPT & \xmark & \underline{0.70 \footnotesize{$\pm$ 0.06}} & 0.63 \footnotesize{$\pm$ 0.04} \\
Semantic entropy & \xmark & \underline{0.70 \footnotesize{$\pm$ 0.05}} & \underline{0.70 \footnotesize{$\pm$ 0.04}} \\
EigenScore & \xmark & 0.68 \footnotesize{$\pm$ 0.04} & 0.67 \footnotesize{$\pm$ 0.04}\\
HaloScope & \cmark & 0.60 \footnotesize{$\pm$ 0.06} & 0.50 \footnotesize{$\pm$ 0.01} \\
LLM-Check & \cmark & 0.48 \footnotesize{$\pm$ 0.03}  & 0.56 \footnotesize{$\pm$ 0.04} \\
Perplexity & \cmark &0.55 \footnotesize{$\pm$ 0.06} & 0.49 \footnotesize{$\pm$ 0.04} \\
Max entropy & \cmark & 0.62 \footnotesize{$\pm$ 0.04} & 0.69 \footnotesize{$\pm$ 0.05} \\
ReDEEP & \cmark & 0.49 \footnotesize{$\pm$ 0.04}  & 0.62 \footnotesize{$\pm$ 0.05} \\
TOHA (ours) & \cmark & \textbf{0.71 \footnotesize{$\pm$ 0.08}} & \textbf{0.80 \footnotesize{$\pm$ 0.03}} \\ 
\hline
\end{tabular}}
\label{tab:hotpot_qa}
\end{table}

\subsection{Analysis of Hallucination-Aware Attention Heads}

\begin{figure}[t]
   \includegraphics[width=\linewidth]{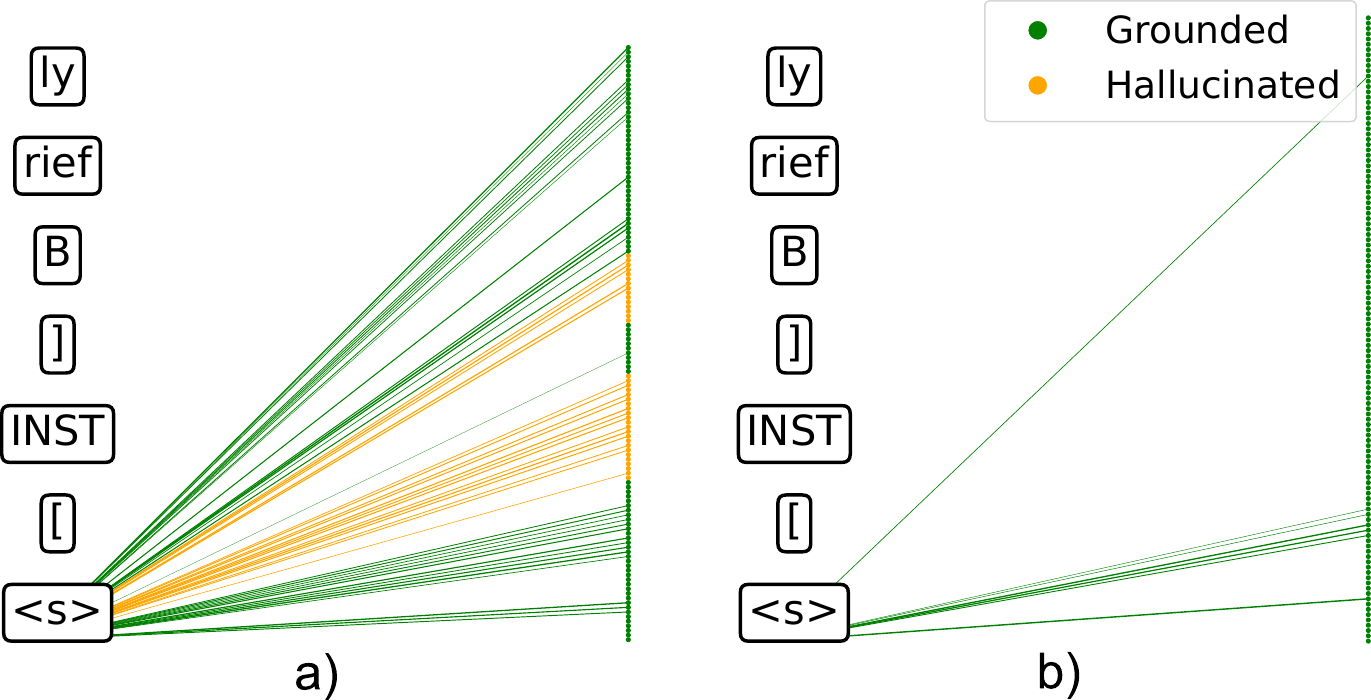} 
    \caption{Attention to the first token (\texttt{<s>} in this example) for (a) a hallucinated generation and (b) a grounded one. Green highlights edges and nodes corresponding to grounded tokens, while yellow indicates hallucinated tokens. Model: Mistral-7B.} \label{fig:bos_pattern}
\end{figure}

\begin{figure}
   \includegraphics[width=\linewidth]{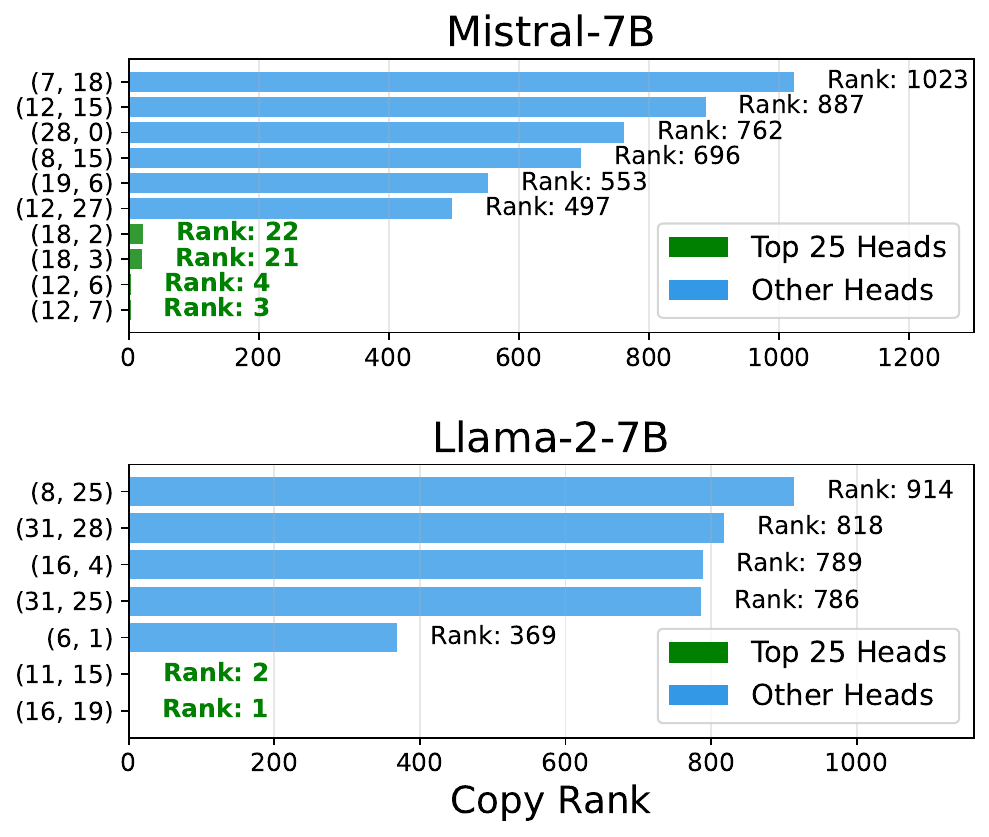} 
    \caption{Copying ranks of hallucination-aware attention heads (lower ranks indicate stronger copying behavior). The row labels $(X, Y)$ correspond to $X$-th layer and $Y$-th head.} \label{fig:copying}
\end{figure}

\paragraph{Attention patterns and copying behavior.} 
For hallucination-aware heads, we analyzed MSF's patterns that distinguish hallucinated from grounded samples. A key finding is that for these heads, hallucinated samples frequently exhibit strong attention to the first token, whereas grounded samples tend to attend to the first token less. An example is provided in Figure~\ref{fig:bos_pattern}. Since this pattern is a known behavior of \textit{induction}, or \textit{token copying} heads~\cite{feucht2025dual}~---~which default to the first token when unable to find previous occurrences of the current token pattern in the context~\cite{elhage2021mathematical}~---~we decided to explore the relationship between these special heads and hallucination-aware ones. We ranked all attention heads in LLama-2-7B and Mistral-7B based on their token copying (induction) scores, following the method of~\cite{feucht2025dual}. For the subset of heads frequently identified by TOHA (appearing in $\geq 20\%$ of runs across all datasets), their copying ranks were recorded. The results (Figure~\ref{fig:copying}) reveal that hallucination-aware heads are often also among the model's top-25 copiers, a finding that aligns with prior work~\cite{sun2025redeep}.

\subsection{Ablation Studies}\label{sec:ablation}

\paragraph{Why zero out the distances between prompt tokens?} 
We hypothesize that the connections within the prompt contribute little to hallucination detection; therefore, our method is designed to disregard them. To validate this architectural choice, we considered an ablation study, where $\operatorname{MTop-Div}_G(R, P)$ was replaced by the MST length of the complete graph in Algorithm~\ref{alg:toha}. The results showing the effectiveness of the proposed approach are in Table~\ref{tab:full_vs_quotient}. 

\begin{table}
\centering
\caption{ROC-AUC $(\uparrow)$ of Algorithm~\ref{alg:toha} with MST length of the complete graph vs $\operatorname{MTop-Div}_G(R, P)$. Best results are highlighted in \textbf{bold}. }
 \label{tab:full_vs_quotient}
\resizebox{\columnwidth}{!}{\begin{tabular}{lcc}
\hline
Dataset & MST length & $\operatorname{MTop-Div}_G(R, P)$ \\
\hline
\multicolumn{3}{c}{Mistral-7B} \\
\hline
CoQA & $0.75 \pm 0.03$ & $\mathbf{0.90 \pm 0.01}$ \\
MS MARCO & $0.38 \pm 0.03$ & $\mathbf{0.65 \pm 0.02}$ \\
\hline
\multicolumn{3}{c}{LLama-2-7B} \\
\hline
CoQA & $0.60 \pm 0.03$ & $\mathbf{0.89 \pm 0.01}$ \\
MS MARCO & $0.37 \pm 0.02$ & $\mathbf{0.76 \pm 0.04}$ \\
\hline
\end{tabular}}
\end{table}

\paragraph{What about other attention map-based features?} To demonstrate that our topology-based approach offers an advantage over conventional attention map features, we compared them to $\operatorname{MTop-Div}$ in the supervised setting. Table~\ref{tab:naive_supervised} (Appendix~\ref{sec:appendix_other_results}) shows that a classifier trained on $\operatorname{MTop-Div}_G(R, P)$ achieves the best performance, confirming that our score captures unique patterns in an attention map that standard approaches miss.

\paragraph{How large should the probe sets be?}
To evaluate TOHA's sensitivity to probe set size, we conducted a sensitivity study. The results shown in Figure~\ref{fig:probe_set_size_ablation} confirm TOHA's robustness: even with only $50$ samples, performance does not drop significantly and remains mostly stable as the probe set size increases. 

\paragraph{Can TOHA operate without any annotated data?} As demonstrated in Section 4.3, many of the attention heads naturally selected by TOHA exhibit strong inductive (token copying) behavior. Crucially, identifying these heads does not require any labeled hallucination data \cite{feucht2025dual}. To evaluate TOHA in an extreme zero-shot setting, we completely bypassed the probe set selection and simply averaged the divergence scores from the model's top-4 copying heads ($\text{TOHA}_{\text{Copy Heads}}$). 

The results in Table \ref{tab:zero_shot_copy} confirm that this fully unsupervised variant is highly competitive, demonstrating that TOHA can be effectively deployed even when zero annotated data is available.

\begin{table*}[t]
\centering
\resizebox{0.9\textwidth}{!}{
\begin{tabular}{lccccc}
\toprule
\multirow{2}{*}{Method} & \multirow{2}{*}{\begin{tabular}[c]{@{}c@{}}Single \\ generation\end{tabular}} & \multicolumn{2}{c}{LLaMA-2-7B} & \multicolumn{2}{c}{Mistral-7B} \\
\cmidrule(lr){3-4} \cmidrule(lr){5-6}
 & & CoQA & MS MARCO & CoQA & MS MARCO \\
\midrule
SelfCheckGPT [1] & \ding{55} & 0.66 $\pm$ 0.03 & 0.59 $\pm$ 0.03 & \textbf{0.86 $\pm$ 0.02} & 0.63 $\pm$ 0.04 \\
Semantic entropy [2] & \ding{55} & \underline{0.76 $\pm$ 0.01} & 0.53 $\pm$ 0.03 & \underline{0.83 $\pm$ 0.02} & 0.54 $\pm$ 0.03 \\
EigenScore [3] & \ding{55} & 0.61 $\pm$ 0.03 & 0.55 $\pm$ 0.03 & 0.74 $\pm$ 0.02 & 0.54 $\pm$ 0.04 \\
HaloScope [4] & \checkmark & 0.61 $\pm$ 0.04 & 0.51 $\pm$ 0.05 & 0.62 $\pm$ 0.08 & 0.57 $\pm$ 0.08 \\
LLM-Check [5] & \checkmark & 0.60 $\pm$ 0.03 & 0.44 $\pm$ 0.02 & 0.60 $\pm$ 0.01 & 0.49 $\pm$ 0.03 \\
Perplexity [6] & \checkmark & 0.74 $\pm$ 0.02 & 0.54 $\pm$ 0.04 & 0.54 $\pm$ 0.03 & 0.45 $\pm$ 0.01 \\
Max entropy [7] & \checkmark & 0.65 $\pm$ 0.03 & \textbf{0.65 $\pm$ 0.04} & 0.73 $\pm$ 0.04 & \textbf{0.68 $\pm$ 0.04} \\
ReDeEP [8] & \checkmark & 0.72 $\pm$ 0.04 & 0.54 $\pm$ 0.04 & 0.59 $\pm$ 0.03 & 0.54 $\pm$ 0.02 \\
$\text{TOHA}_{\text{Copy Heads}} \,\text{(ours)}$ & \checkmark & \textbf{0.89 $\pm$ 0.01} & \underline{0.62 $\pm$ 0.04} & 0.77 $\pm$ 0.01 & \underline{0.66 $\pm$ 0.04} \\
\bottomrule
\end{tabular}
}
\caption{ROC AUC ($\uparrow$) evaluation of TOHA using only the top-4 copying heads (zero annotated data) compared to baselines. The best results in each column are highlighted in \textbf{bold}, and the second best are \underline{underlined}. }
\label{tab:zero_shot_copy}
\end{table*}

\paragraph{How many attention heads do we need?}
To evaluate the sensitivity of our method to the hyperparameter $N_{max}$, we performed an ablation study for values from $1$ to $10$. As shown in Figure~\ref{fig:n_max_comparison}, TOHA achieves strong detection performance even when $N_{\max} = 1$, which underscores the effectiveness of our topological approach.

\section{Related Works}

Existing hallucination detection methods face strict trade-offs~\cite{zhang2023siren, huang2023survey, wang2024factuality}. Consistency-based approaches~\cite{manakul2023selfcheckgpt, chen2024inside, qiusemantic, nikitin2024kernel, hou2025probabilistic} are robust but computationally expensive due to multiple generation passes. Surface-level metrics like perplexity~\cite{fadeeva2024fact, malinin2021uncertainty} are efficient yet limited, as they ignore the model's rich internal representations~\cite{gekhman2025insideout}. Conversely, hidden-state classifiers~\cite{azaria-mitchell-2023-internal, sky2024androids, zhou2025hademif} capture these representations but require extensive annotated data and suffer from poor task transferability~\cite{sky2024androids}. Even semi-supervised alternatives like HaloScope~\cite{du2024haloscope} still demand large volumes of unannotated outputs.

Attention map-based methods represent a promising yet underdeveloped direction. Current techniques either (i) rely on large annotated datasets~\cite{chuang2024lookbacklensdetectingmitigating, binkowski2025hallucination}, (ii) exploit only simplistic attention graph properties like self-loop weights~\cite{sriramanan2024llmcheck}, or (iii) disregard attention map geometry entirely, using mechanistic scores instead~\cite{sun2025redeep}. 
Thus, training-free methods that fully leverage the rich structural information encoded in attention graphs remain underexplored.

\section{Conclusion}

This paper introduces TOHA, a novel hallucination detection method based on the topological structure of attention maps. Central to TOHA is the $\operatorname{MTop-Div}_G(R, P)$, our adaptation of Manifold Topology Divergence~\cite{barannikov2021manifold} for the graph setting, which, as we demonstrate, serves as a measure of the response novelty in relation to the prompt. This property makes the proposed divergence well-suited for hallucination detection in RAG scenarios. 

By analyzing divergence values, we identified a subset of ``hallucination-aware'' attention heads that reliably distinguish hallucinated from grounded samples across datasets. TOHA computes the final hallucination scores by averaging the topological divergences from these heads. Further investigation reveals that some of these heads are associated with copying behavior, which aligns with prior work~\cite{sun2025redeep}.

Extensive experiments show that TOHA is a robust alternative to existing approaches, matching or surpassing state-of-the-art baselines. Moreover, our method is both data- and compute-efficient: just $50$ annotated samples suffice for reliable detection, and inference runs several times faster than comparable methods of similar quality. Crucially, we validate TOHA's transferability, demonstrating its robustness to shifts in data distribution~---~a key advantage for real-world deployment, where LLM inputs are far more diverse and complex than benchmark examples.

In summary, TOHA delivers state-of-the-art detection performance while combining efficiency and solid generalizability, making it particularly well-suited for practical applications.

\section*{Acknowledgements}

The authors wish to thank our colleague Ilya Kuleshov for his time and the many insightful discussions regarding this research. The work was supported by the grant for research centers in the field of AI provided by the Ministry of Economic Development of the Russian Federation in accordance with the agreement 000000C313925P4F0002 and the agreement №139-10-2025-033.

\section*{Limitations}

While TOHA demonstrates strong performance and efficiency, several limitations warrant discussion. 

\textbf{Model-specific dependencies.} TOHA’s effectiveness relies on identifying ``hallucination-aware'' attention heads, which may vary across LLM architectures. While our experiments cover popular open-source models (e.g., LLaMA, Mistral), further validation is needed for proprietary or larger models (e.g., GPT-4, Claude). 

\textbf{Multimodal extensions.} The current framework operates solely on text. Adapting TOHA to multimodal settings (e.g., vision-language models) would require redefining attention graphs across heterogeneous data modalities.  

\newpage

\bibliography{example_paper}

\appendix

\section{Additional Experimental Results}\label{sec:appendix_other_results}

\subsection{Multiple Generations-based Methods}

To provide a complete comparison, we considered an implementation of consistency-based methods with $20$ additional generations. The results are demonstrated in Tables~\ref{tab:additional_generations}-\ref{tab:additional_generations_newer_models}. We can see that TOHA still remains superior to the baselines, achieving top performance in most experiments. 

\begin{table*}[h!]
\centering
\caption{ROC AUC $(\uparrow)$ of multiple generations-based methods (with 20 additional samples) and TOHA. The best results for each model are highlighted in \textbf{bold}, and the second best are \underline{underlined}.}
 \label{tab:additional_generations}
 \renewcommand{\arraystretch}{1} 
\resizebox{0.9\textwidth}{!}{%
\begin{tabular}{lccccccc}
\hline
\multirow{2}{*}{Method} & Single &\multirow{2}{*}{MS MARCO} & CNN/DM + &  \multirow{2}{*}{CoQA} &  \multirow{2}{*}{SQuAD} &  \multirow{2}{*} {XSum} \\ 
 & generation & & Recent News &  & & & \\ \hline
\multicolumn{7}{c}{Mistral-7B} \\ \hline
SelfCheckGPT~\citetalias{manakul2023selfcheckgpt} & \xmark & \underline{0.67 \footnotesize{$\pm$ 0.03}} & \underline{0.59 \footnotesize{$\pm$ 0.04}}  & \textbf{0.93 \footnotesize{$\pm$ 0.01}} & \underline{0.83 \footnotesize{$\pm$ 0.02}} & \textbf{0.71 \footnotesize{$\pm$ 0.04}} \\
Semantic entropy~\citetalias{farquhar2024detecting} & \xmark & 0.53 \footnotesize{$\pm$ 0.03} & 0.52 \footnotesize{$\pm$ 0.03} & 0.86 \footnotesize{$\pm$ 0.01} & 0.74 \footnotesize{$\pm$ 0.01} & 0.63 \footnotesize{$\pm$ 0.02} \\
EigenScore~\citetalias{chen2024inside} & \xmark & 0.49 \footnotesize{$\pm$ 0.03} & 0.53 \footnotesize{$\pm$ 0.05} & 0.78 \footnotesize{$\pm$ 0.02} & 0.77 \footnotesize{$\pm$ 0.04} & 0.59 \footnotesize{$\pm$ 0.05} \\
TOHA (ours) & \cmark & \textbf{0.76 \footnotesize{$\pm$ 0.04}} & 
\textbf{0.60 \footnotesize{$\pm$ 0.09}} & \underline{0.89 \footnotesize{$\pm$ 0.01}} & \textbf{0.96 \footnotesize{$\pm$ 0.01}} & \underline{0.66 $\pm$ 0.05} \\ \hline
\multicolumn{7}{c}{LLama-2-7B} \\ \hline
SelfCheckGPT~\citetalias{manakul2023selfcheckgpt} & \xmark & \underline{0.60 \footnotesize{$\pm$ 0.04}} & \textbf{0.60 \footnotesize{$\pm$ 0.03}}  & 0.77 \footnotesize{$\pm$ 0.02} & \underline{0.78 \footnotesize{$\pm$ 0.02}} & \underline{0.67 \footnotesize{$\pm$ 0.04}} \\
Semantic entropy~\citetalias{farquhar2024detecting} & \xmark & 0.56 \footnotesize{$\pm$ 0.03} & 0.49 \footnotesize{$\pm$ 0.03} & \underline{0.79 \footnotesize{$\pm$ 0.01}} & 0.77 \footnotesize{$\pm$ 0.02} & 0.63 \footnotesize{$\pm$ 0.04} \\
EigenScore~\citetalias{chen2024inside} & \xmark & 0.57 \footnotesize{$\pm$ 0.04} & 0.52 \footnotesize{$\pm$ 0.06} & 0.61 \footnotesize{$\pm$ 0.03} & \underline{0.78 \footnotesize{$\pm$ 0.02}} & 0.65 \footnotesize{$\pm$ 0.03} \\
TOHA (ours) & \cmark & \textbf{0.65 \footnotesize{$\pm$ 0.02}} & \underline{0.56 \footnotesize{$\pm$ 0.02}} & \textbf{0.90 \footnotesize{$\pm$ 0.01}} & \textbf{0.87 \footnotesize{$\pm$ 0.04}} & \textbf{0.68 \footnotesize{$\pm$ 0.05}} \\ \hline
\multicolumn{7}{c}{LLaMA-2-13B} \\ \hline
SelfCheckGPT~\citetalias{manakul2023selfcheckgpt} & \xmark & \underline{0.61 \footnotesize{$\pm$ 0.05}} & \textbf{0.60 \footnotesize{$\pm$ 0.06}}  & \underline{0.88 \footnotesize{$\pm$ 0.02}} & \underline{0.75 \footnotesize{$\pm$ 0.04}} & 0.61 \footnotesize{$\pm$ 0.04} \\
Semantic entropy~\citetalias{farquhar2024detecting} & \xmark & 0.60 \footnotesize{$\pm$ 0.03} & 0.52 \footnotesize{$\pm$ 0.03} & 0.77 \footnotesize{$\pm$ 0.04} & 0.70 \footnotesize{$\pm$ 0.02} & \underline{0.62 \footnotesize{$\pm$ 0.03}} \\
EigenScore~\citetalias{chen2024inside} & \xmark & 0.58 \footnotesize{$\pm$ 0.04} & 0.48 \footnotesize{$\pm$ 0.05} & 0.59 \footnotesize{$\pm$ 0.03} & 0.60 \footnotesize{$\pm$ 0.02} & 0.54 \footnotesize{$\pm$ 0.05} \\
TOHA (ours) & \cmark & \textbf{0.67 \footnotesize{$\pm$ 0.04}} & \underline{0.56 \footnotesize{$\pm$ 0.05}} & \textbf{0.92 \footnotesize{$\pm$ 0.02}} & \textbf{0.88 \footnotesize{$\pm$ 0.05}} & \textbf{0.66 \footnotesize{$\pm$ 0.03}} \\
 \hline
\end{tabular}}
\end{table*}

\begin{table}[t!]
\centering
\caption{ROC AUC $(\uparrow)$ of multiple generations-based methods (with 20 additional samples) and TOHA. The best results for each model are highlighted in \textbf{bold}, and the second best are \underline{underlined}.}
 \label{tab:additional_generations_newer_models}
\resizebox{\columnwidth}{!}{\begin{tabular}{lcccc}
\hline
Method & Single gen. & CoQA & SQuAD & XSum \\ \hline
\multicolumn{5}{c}{LLaMA-3.1-8B} \\ \hline
SelfCheckGPT~\citetalias{manakul2023selfcheckgpt} & \xmark & \textbf{0.95 \footnotesize{$\pm$ 0.01}} & \underline{0.78 \footnotesize{$\pm$ 0.03}} & \textbf{0.75 \footnotesize{$\pm$ 0.03}} \\
Semantic entropy~\citetalias{farquhar2024detecting} & \xmark & 0.82 \footnotesize{$\pm$ 0.03} & 0.54 \footnotesize{$\pm$ 0.06} & 0.46 \footnotesize{$\pm$ 0.05} \\
EigenScore~\citetalias{chen2024inside} & \xmark & \underline{0.84 \footnotesize{$\pm$ 0.01}} & 0.56 \footnotesize{$\pm$ 0.03} & 0.48 \footnotesize{$\pm$ 0.03} \\
TOHA (ours) & \cmark & \underline{0.84 \footnotesize{$\pm$ 0.01}} & \textbf{0.87} \footnotesize{$\pm$ 0.03} & \underline{0.65 \footnotesize{$\pm$ 0.05}} \\ \hline
\multicolumn{5}{c}{Qwen2.5-7B} \\ \hline
SelfCheckGPT~\citetalias{manakul2023selfcheckgpt} & \xmark & 0.75 \footnotesize{$\pm$ 0.02}  & \textbf{0.77 \footnotesize{$\pm$ 0.02}} & \textbf{0.72 \footnotesize{$\pm$ 0.03}}  \\
Semantic entropy~\citetalias{farquhar2024detecting} & \xmark & 0.83 \footnotesize{$\pm$ 0.05} &  \underline{0.58 \footnotesize{$\pm$ 0.04}} & \underline{0.69 \footnotesize{$\pm$ 0.05}}\\
EigenScore~\citetalias{chen2024inside} & \xmark & 0.83 \footnotesize{$\pm$ 0.05} & 0.56 \footnotesize{$\pm$ 0.03}  & 0.48 \footnotesize{$\pm$ 0.03}\\
TOHA (ours)&  \cmark & \underline{0.79 \footnotesize{$\pm$ 0.05}} & \textbf{0.77 \footnotesize{$\pm$ 0.02}}  & \underline{0.69 \footnotesize{$\pm$ 0.03}}\\ \hline
\end{tabular}}
\end{table}

\subsection{Comparison with ReDeEP}

To provide a comprehensive comparison with ReDeEP~\cite{sun2025redeep}, which was evaluated on the entire RAGTruth dataset in the original paper, we conducted a similar evaluation of TOHA under the same conditions. As shown in Table~\ref{tab:ragtruth_full}, TOHA not only outperforms ReDeEP but also demonstrates greater robustness to data distribution shifts, exhibiting less significant performance degradation on separate benchmarks of RAGTruth (Tables~\ref{tab:mistral}-\ref{tab:llama_3_8b}).

\begin{table}[h]
\centering
\caption{Performance comparison between ReDEEP and TOHA on the entire RAGTruth dataset.}
 \renewcommand{\arraystretch}{1} 
\resizebox{0.85\columnwidth}{!}{%
\begin{tabular}{lcc}
\hline
Method & Llama-2-7B & Llama-2-13B \\
\hline
ReDeEP~\citetalias{sun2025redeep} &  0.68 \footnotesize{$\pm$ 0.02} & 0.77 \footnotesize{$\pm$ 0.01} \\
TOHA (ours) &  \textbf{0.70 $\pm$ \footnotesize{0.04}}  & \textbf{0.80 $\pm$ \footnotesize{0.02}} \\
\hline
\end{tabular}}
\label{tab:ragtruth_full}
\end{table}

\subsection{Alternative Attention-based Features for Hallucination Detection}

\begin{table}[h!]
\caption{ROC-AUC values of supervised classifiers on top of various sets of features. TOP-1 results are highlighted with \textbf{bold font}, while TOP-2 are \underline{underlined}.}
 \label{tab:naive_supervised}
 \centering
\resizebox{0.85\columnwidth}{!}{%
\begin{tabular}{lcc}
\hline
Features & MS MARCO  & CoQA \\ \hline
\multicolumn{3}{c}{Mistral-7B} \\ \hline
Standard topological & 0.67 & 0.69 \\
Sparsity ratio & 0.66 & 0.7 \\ 
Entropy & 0.75 & \underline{0.77} \\
Wasserstein & \underline{0.77} & 0.73 \\
Spectral norm & 0.73 & 0.72 \\
Attention to \texttt{<s>} & 0.65 & 0.61 \\
MTop-Div & \textbf{0.86} & \textbf{0.98} \\ \hline
\multicolumn{3}{c}{LLaMA-2-7B} \\ \hline
Standard topological & 0.69 & 0.7 \\
Sparsity ratio & 0.49 & 0.61 \\ 
Entropy & 0.38 & \underline{0.68} \\
Wasserstein & \underline{0.73} & 0.6 \\
Spectral norm & 0.49 & 0.64 \\
Attention to \texttt{<s>} & 0.62 & 0.64 \\
MTop-Div & \textbf{0.75} & \textbf{0.96} \\ \hline
\end{tabular}}
\end{table}

During preliminary experiments for an attention map-based hallucination detector, we evaluated a range of topological and traditional features. As standard topological features, we employed barcode-based features, such as the sum of bar lengths in persistence diagrams, and naive topological features, including the average vertex degree in attention graphs~\cite{kushnareva2021artificial}. As for traditional attention-based features, we used sparsity ratio, attention entropy, and spectral norm~\cite{kobayashi2020attention,vig-belinkov-2019-analyzing, ji-etal-2021-distribution}. We also considered Wasserstein distances between the persistent diagrams~\cite{chazal2017introduction} of the prompt and response subgraphs as an alternative measure of their similarity.  Finally, we analyzed average attention to the first token and found that hallucination-aware heads often attend in the presence of hallucinations (see Section~\ref{sec:experiments}).

To identify the most informative features for hallucination detection, we trained $L1$-regularized supervised classifiers on concatenated features from all layers and heads and compared them to classifiers trained only on $\operatorname{MTop-Div}_G(R, P)$ values under the same conditions. The results are presented in Table~\ref{tab:naive_supervised}.

While the classifier based on $\operatorname{MTop-Div}_G(R, P)$ values significantly outperforms alternative approaches, computing these values across all layers and attention heads is highly computationally expensive. To address this, we developed TOHA~---~a more efficient alternative that aggregates divergence values from only a subset of the ``hallucination-aware'' attention heads.

\subsection{Evaluation of Alternative Head Selection Metrics}

\begin{figure*}[t!]
    \centering
    \includegraphics[width=\textwidth]{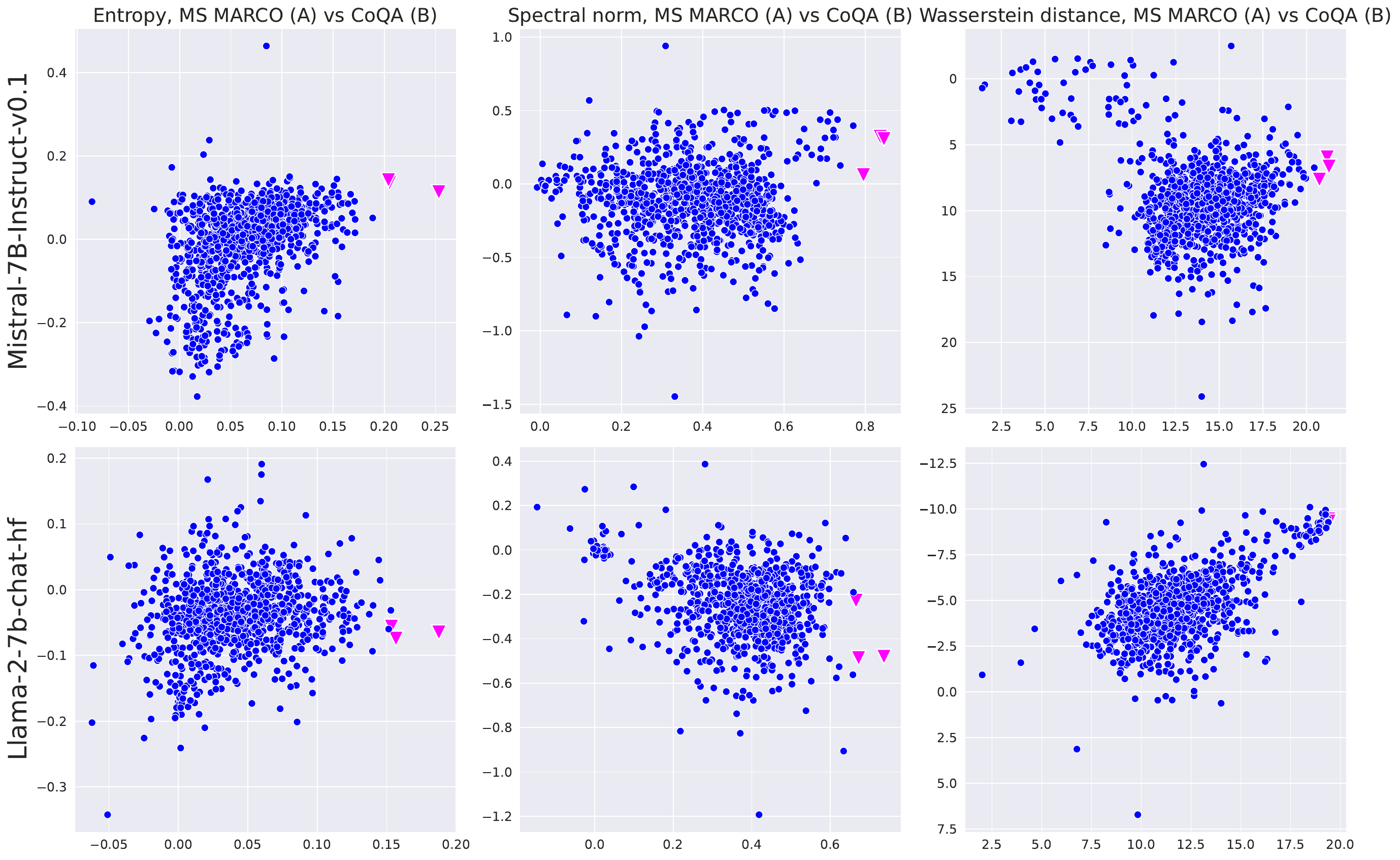} %
    \caption{$\Delta_{ij}$ values for $ij$ heads, MS MARCO and CoQA. Vertical axis corresponds to the difference in the dataset (B), horizontal axis to the difference in the dataset (A). The heads that separate samples best are highlighted in pink.} \label{fig:average_distance_alternative}
\end{figure*}

Additionally, we investigated alternative attention map-based scores, such as entropy, spectral norm, and the Wasserstein distance between the persistent diagrams of prompts and responses, for selecting specialized attention heads. Following the pipeline of the Algorithm~\ref{alg:toha}, we computed the average distances between hallucinated and grounded samples using alternative scores. The results, presented in Figure~\ref{fig:average_distance_alternative}, reveal that the heads that separate samples best for MS MARCO do not generalize to the CoQA dataset. This suggests that our proposed $\operatorname{MTop-Div}_G(R, P)$ metric is better suited for the task 
compared to existing solutions.

\subsection{Feature-Level Comparison with LookBack Lens}
To further validate the representational power of our proposed $\operatorname{MTop-Div}_G(R, P)$  features, we compared them against the attention-ratio features introduced in Lookback Lens~\cite{chuang2024lookbacklensdetectingmitigating}. As Lookback Lens is fundamentally a supervised method, we established a fair, comparable setting by isolating its attention-ratio features and evaluating them using the same attention-head selection procedure as TOHA. The results of this comparison are provided in Table~\ref{tab:attention_ratio_comparison}. TOHA consistently outperforms the attention-ratio features across all evaluated models and datasets, confirming the superior effectiveness of topological features over standard attention metrics for hallucination detection.
\begin{table}[htbp]
    \centering
    \small
    \caption{Performance comparison between TOHA and attention-ratio features (derived from LookBack Lens) evaluated under identical settings.}
    \label{tab:attention_ratio_comparison}
    \renewcommand{\arraystretch}{1} 
\resizebox{0.9\columnwidth}{!}{
\begin{tabular}{lcc}
        \toprule
        Method & CoQA & XSum \\
        \midrule
        \multicolumn{3}{c}{Llama-2-7B} \\
        \midrule
        Attention ratio & 0.76 $\pm$ 0.05 & 0.59 $\pm$ 0.05 \\
        TOHA (ours)     & \textbf{0.90 $\pm$ 0.01} & \textbf{0.68 $\pm$ 0.05} \\
        \midrule
        \multicolumn{3}{c}{Qwen2.5-7B} \\
        \midrule
        Attention ratio & 0.71 $\pm$ 0.07 & 0.58 $\pm$ 0.04 \\
        TOHA (ours)     & \textbf{0.79 $\pm$ 0.05} & \textbf{0.69 $\pm$ 0.03} \\
        \bottomrule
    \end{tabular}}
\end{table}

\subsection{Comparison with Linear Probes under Matched Data Constraints}
\label{sec:linear_probes}

To ensure a fair comparison with supervised baseline approaches, we evaluated the performance of linear probes trained on datasets with exactly 100 samples---matching the probe set size used by TOHA. We conducted these experiments across all five datasets (MS MARCO, CNN/DM, CoQA, SQuAD, and XSum) using the LLaMA-2-13B model. 

The corresponding results are detailed in Table \ref{tab:linear_probes}. While linear probes achieve the second-best performance in two out of the five experiments, they are significantly outperformed by TOHA across the board. This highlights the superior sample efficiency and robustness of our topology-based method when operating under constrained data availability.

\begin{table*}[t]
\centering
\caption{ROC AUC ($\uparrow$) evaluation of a linear probe trained on exactly 100 samples alongside TOHA and other baseline methods. The best results for the model are highlighted in \textbf{bold}, and the second best are \underline{underlined}.}
\resizebox{0.9\textwidth}{!}{
\begin{tabular}{lcccccc}
\toprule
Method & \begin{tabular}[c]{@{}c@{}}Single \\ generation\end{tabular} & MS MARCO & \begin{tabular}[c]{@{}c@{}}CNN/DM + \\ Recent News\end{tabular} & CoQA & SQuAD & XSum \\
\midrule
\multicolumn{7}{c}{LLaMA-2-13B} \\
\midrule
SelfCheckGPT [1] & \ding{55} & 0.58 $\pm$ 0.04 & \textbf{0.58 $\pm$ 0.05} & \underline{0.77 $\pm$ 0.02} & 0.64 $\pm$ 0.03 & \underline{0.60 $\pm$ 0.04} \\
Semantic entropy [2] & \ding{55} & 0.57 $\pm$ 0.04 & 0.54 $\pm$ 0.03 & 0.76 $\pm$ 0.04 & 0.65 $\pm$ 0.03 & \underline{0.60 $\pm$ 0.03} \\
EigenScore [3] & \ding{55} & 0.56 $\pm$ 0.04 & 0.47 $\pm$ 0.04 & 0.57 $\pm$ 0.03 & 0.57 $\pm$ 0.02 & 0.52 $\pm$ 0.06 \\
HaloScope [4] & \checkmark & 0.54 $\pm$ 0.09 & 0.51 $\pm$ 0.04 & 0.57 $\pm$ 0.03 & 0.55 $\pm$ 0.02 & 0.55 $\pm$ 0.07 \\
LLM-Check [5] & \checkmark & 0.49 $\pm$ 0.06 & \underline{0.56 $\pm$ 0.05} & 0.57 $\pm$ 0.05 & 0.57 $\pm$ 0.07 & 0.57 $\pm$ 0.07 \\
Perplexity [6] & \checkmark & 0.54 $\pm$ 0.04 & 0.46 $\pm$ 0.07 & 0.62 $\pm$ 0.03 & 0.45 $\pm$ 0.02 & 0.49 $\pm$ 0.05 \\
Max entropy [7] & \checkmark & \underline{0.62 $\pm$ 0.03} & 0.53 $\pm$ 0.06 & 0.66 $\pm$ 0.03 & \underline{0.78 $\pm$ 0.02} & 0.59 $\pm$ 0.04 \\
ReDEEP [8] & \checkmark & \underline{0.62 $\pm$ 0.06} & 0.48 $\pm$ 0.05 & 0.73 $\pm$ 0.02 & 0.48 $\pm$ 0.07 & 0.58 $\pm$ 0.08 \\
Linear probe & \checkmark & \underline{0.62 $\pm$ 0.03} & 0.55 $\pm$ 0.05 & 0.65 $\pm$ 0.03 & \underline{0.78 $\pm$ 0.01} & 0.58 $\pm$ 0.05 \\
TOHA (ours) & \checkmark & \textbf{0.67 $\pm$ 0.04} & \underline{0.56 $\pm$ 0.05} & \textbf{0.92 $\pm$ 0.02} & \textbf{0.88 $\pm$ 0.05} & \textbf{0.66 $\pm$ 0.03} \\
\bottomrule
\end{tabular}
}
\label{tab:linear_probes}
\end{table*}

\subsection{Performance Analysis on Out-of-Design Cases}

\begin{figure}[t!]
    \centering
    \includegraphics[width=\columnwidth]{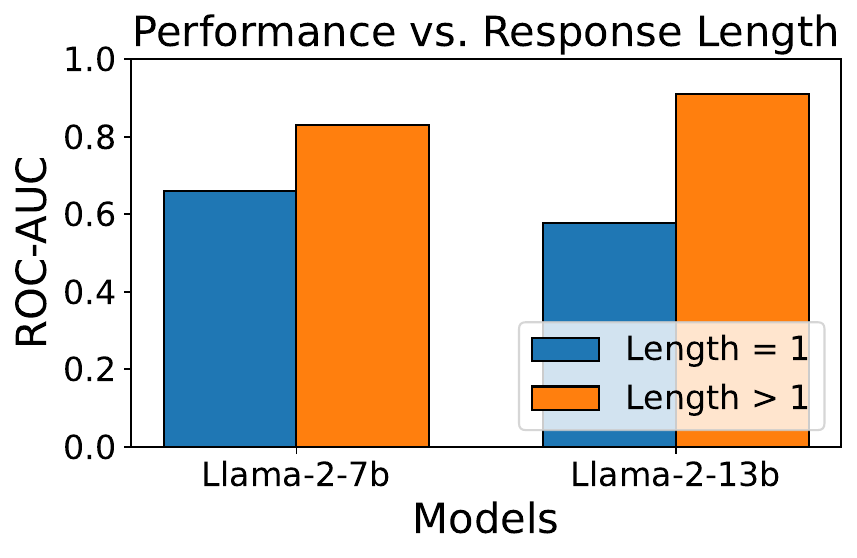} 
    \caption{Comparison of ROC-AUC scores ($\uparrow$) for single-word versus multi-word model responses. Dataset: SQuAD.} \label{fig:short_answers}
\end{figure}

Additionally, we analyzed TOHA's performance on extremely short responses consisting of a single word. The results are presented in Figure~\ref{fig:short_answers}. While its performance drops, as expected when the response graph degenerates to a single vertex, it remains non-random. This indicates that TOHA retains some predictive power even for these challenging, out-of-design cases. 

As for the TOHA's performance on longer responses, the results for the MS MARCO, CNN/DM, and XSum datasets demonstrate that as the response length grows, the hallucination signals become less distinguishable for all the considered methods, as hallucinated tokens comprise a very small part of the response. However, TOHA remains superior to baselines, confirming that our topology-based approach is more effective than existing methods for complex long-form responses.

\subsection{Statistical Significance of the Results}

To rigorously evaluate the robustness and consistency of our results, we conducted a statistical significance analysis of the performance differences across all experimental settings. Specifically, we applied a non-parametric Wilcoxon signed-rank test followed by Holm's step-down procedure to correct for multiple comparisons. The outcomes of this post-hoc analysis are presented in Tables~\ref{tab:pairwise_pvalues} and \ref{tab:ranking}. 

As detailed in Table~\ref{tab:pairwise_pvalues}, the pairwise $p$-values confirm that TOHA yields a statistically significant difference against every single baseline evaluated. Furthermore, the aggregate rankings in Table~\ref{tab:ranking} show that TOHA achieves the best overall average rank (1.67) and stands alone at the top of the hierarchy, clearly distinguishable from all other methods.

\begin{table}[htbp]
    \centering
    \caption{Pairwise $p$-values from the Wilcoxon-Holm post-hoc analysis. Methods are considered statistically different after thresholding the $p$-value by the normalized Holm correction. Notice that TOHA yields a ``True'' significance against every single baseline evaluated.}
    \small 
    \begin{tabular}{llcc}
        \toprule
        Method 1 & Method 2 & $p$-value & Stat. diff. \\
        \midrule
        Haloscope & TOHA & $9.5 \times 10^{-7}$ & True \\
        LLM-Check & SelfCK & $9.5 \times 10^{-7}$ & True \\
        LLM-Check & TOHA & $9.5 \times 10^{-7}$ & True \\
        ReDEEP & TOHA & $9.5 \times 10^{-7}$ & True \\
        Sem. Entropy & TOHA & $9.5 \times 10^{-7}$ & True \\
        EigenScore & TOHA & $1.9 \times 10^{-6}$ & True \\
        Perplexity & TOHA & $1.9 \times 10^{-6}$ & True \\
        Entropy & LLM-Check & $5.2 \times 10^{-5}$ & True \\
        ReDEEP & SelfCK & $6.7 \times 10^{-5}$ & True \\
        LLM-Check & Sem. Entropy & $1.0 \times 10^{-4}$ & True \\
        Perplexity & SelfCK & $2.9 \times 10^{-4}$ & True \\
        Entropy & TOHA & $1.0 \times 10^{-3}$ & True \\
        EigenScore & Entropy & $1.6 \times 10^{-3}$ & True \\
        SelfCK & TOHA & $1.6 \times 10^{-3}$ & True \\
        \midrule 
        Entropy & ReDeEP & $2.5 \times 10^{-3}$ & False \\
        Haloscope & LLM-Check & $2.9 \times 10^{-3}$ & False \\
        EigenScore & SelfCK & $3.8 \times 10^{-3}$ & False \\
        Entropy & Perplexity & $4.9 \times 10^{-3}$ & False \\
        Haloscope & Perplexity & $1.8 \times 10^{-2}$ & False \\
        EigenScore & LLM-Check & $1.9 \times 10^{-2}$ & False \\
        ReDeEP & Sem. Entropy & $2.2 \times 10^{-2}$ & False \\
        SelfCK & Sem. Entropy & $2.2 \times 10^{-2}$ & False \\
        Haloscope & SelfCK & $2.6 \times 10^{-2}$ & False \\
        Perplexity & Sem. Entropy & $4.6 \times 10^{-2}$ & False \\
        Entropy & Haloscope & $5.0 \times 10^{-2}$ & False \\
        Entropy & Sem. Entropy & $1.7 \times 10^{-1}$ & False \\
        EigenScore & Sem. Entropy & $2.6 \times 10^{-1}$ & False \\
        Haloscope & Sem. Entropy & $3.2 \times 10^{-1}$ & False \\
        Haloscope & ReDeEP & $3.4 \times 10^{-1}$ & False \\
        EigenScore & Perplexity & $3.7 \times 10^{-1}$ & False \\
        EigenScore & ReDeEP & $3.9 \times 10^{-1}$ & False \\
        LLM-Check & ReDeEP & $4.1 \times 10^{-1}$ & False \\
        Perplexity & ReDeEP & $4.5 \times 10^{-1}$ & False \\
        LLM-Check & Perplexity & $5.9 \times 10^{-1}$ & False \\
        Entropy & SelfCK & $8.9 \times 10^{-1}$ & False \\
        EigenScore & Haloscope & $1.0 \times 10^{0}$ & False \\
        \bottomrule
    \end{tabular}
    \label{tab:pairwise_pvalues}
\end{table}

\begin{table}[htbp]
    \centering
    \caption{Average rank of each method across evaluated settings and the number of baselines from which it is statistically indistinguishable (derived from the Wilcoxon-Holm post-hoc analysis). Lower values ($\downarrow$) indicate better performance. TOHA achieves the top rank and is statistically distinct from all other evaluated approaches.}
     \renewcommand{\arraystretch}{1} 
\resizebox{0.9\linewidth}{!}{%
    \begin{tabular}{lcc}
        \toprule
        Method & Rank $\downarrow$ & \# Indistinguishable $\downarrow$ \\
        \midrule
        TOHA (Ours) & \textbf{1.67} & \textbf{0} \\
        SelfCK & 3.00 & 3 \\
        Entropy & 3.62 & 3 \\
        Semantic Entropy & 4.95 & 3 \\
        HaloScope & 5.71 & 3 \\
        EigenScore & 5.74 & 3 \\
        ReDEEP & 6.38 & 2 \\
        Perplexity & 6.81 & 2 \\
        LLM-Check & 7.12 & 0 \\
        \bottomrule
    \end{tabular}}
    \label{tab:ranking}
\end{table}

\subsection{T-test Analysis for Transferability Experiments}
As previously noted, TOHA's performance on the CNN/DM and XSum datasets in transfer settings falls within the method's standard deviation. To statistically confirm this observation, we calculated $p$-values using a $t$-test for the means of two independent samples. The results, provided in Table~\ref{tab:pvalues}, fully support our claim: all $p$-values are well above the $0.05$ threshold, confirming that the performance variations in transfer settings are not statistically significant.
\begin{table}[htbp]
    \centering
    \caption{$p$-values of the independent two-sample $t$-test for the transferability experiments.}
    \label{tab:pvalues}
    \begin{tabular}{lcc}
        \toprule
        Probe Set & CNN/DM & XSum \\
        \midrule
        MS MARCO & 0.272 & 1.000 \\
        CNN/DM   & 1.000 & 0.746 \\
        CoQA     & 0.499 & 0.134 \\
        SQuAD    & 0.373 & 0.346 \\
        XSum     & 0.515 & 1.000 \\
        \bottomrule
    \end{tabular}
\end{table}

\section{Topological Data Analysis: Background}\label{sec:appendix_topology}

\paragraph{Simplicial complexes.} A simplicial complex $S$ is a collection of simplices such that every face of a simplex $\sigma \in S$ is also in $S$. 
Simplices are the higher-dimensional generalizations of triangles; a 0-simplex is a vertex, a 1-simplex is an edge, a 2-simplex is a triangle, and so forth. Formally, given a finite set $X$, an $n$-simplex $\sigma$ is an $(n+1)$ subset of $X$. 
Simplicial complexes are fundamental objects in algebraic and combinatorial topology, serving as discrete analogs of topological spaces.

\paragraph{Vietoris-Rips simplicial complex.} The Vietoris-Rips complex $VR_{\varepsilon}(X)$ of a weighted graph $G = (V_G, E_G)$ with distance threshold $\varepsilon > 0$ is defined as follows:
$$\mathrm{VR}_\varepsilon(G) = \left\{ \sigma \subseteq V_G \ \bigg| \ \forall\, v_i,v_j \in \sigma, \ w(e_{ij}) \leq \varepsilon \right\},$$
where $w$ is the edge weight function associated with $G$.

\paragraph{Homology groups.} Homology groups \( H_k \) are invariants used in algebraic topology to study the topological properties of a space. Let $C_k(S)$ denote vector space over $\mathbb{Z}/2\mathbb{Z}$, with the basis consisting of $k$-dimensional simplices of $S$. Elements of $C_k$ are called chains. Formally, homology groups are derived from a chain complex $(C_{\bullet}, \partial_{\bullet})$, which is a sequence of $C_k$ connected by boundary maps $\partial_k$:
\begin{align*}
    C_{\bullet}:  \cdots \rightarrow C_{k + 1} \xrightarrow{\partial_{k+1}} C_k \xrightarrow{\partial_k} \cdots, \, \  \\  \partial_k \circ \partial_{k + 1} = 0.
\end{align*}

The \( k \)-th homology group \( H_k \) is defined as the quotient of the group of \( k \)-cycles (chains whose boundary is zero) by the group of \( k \)-boundaries (chains that are the boundary of a \((k+1)\)-chain). Mathematically, this is expressed as:
$$H_k(S) = Z_k(S) / B_k(S),$$
where $Z_k = \mathrm{ker}\, \partial_k = \{ c \in C_k \, |\, \partial_k(c) = 0 \}$  and $ B_k = \mathrm{im} \,\partial_{k + 1} = \{ \partial_{k + 1}(c) \, | \, c \in C_{k + 1}\}$ is the group of \( k \)-boundaries.  The elements of $H_k(S)$ represent various $k$-dimensional topological
features in $S$. Elements of a basis in $H_k(S)$ correspond to a set of basic topological features.

\paragraph{Filtrations.} A filtration of simplicial complexes $\mathcal{F}$ is a family of nested simplicial complexes:
$$\mathcal{F} : \varnothing \subseteq S_1 \subseteq S_2 \subseteq \dots \subseteq S_n = S,$$
where each $S_k$ is a simplicial complex itself. In practice, the filtrations of simplicial complexes are usually obtained for sequences of increasing thresholds $0 < \varepsilon_1 < \dots < \varepsilon_n$. For example, simplicial complexes $VR_{\varepsilon_i}(X)$ form a filtration
\begin{align*}
    \mathcal{F}_{VR}(X) : & \varnothing \subseteq VR_{\varepsilon_1}(X) \subseteq VR_{\varepsilon_2}(X) \subseteq \dots \\ &\subseteq VR_{\varepsilon_n}(X) = VR(X). 
\end{align*}

\begin{figure}[ht!]
    \centering
    \includegraphics[width=\columnwidth]{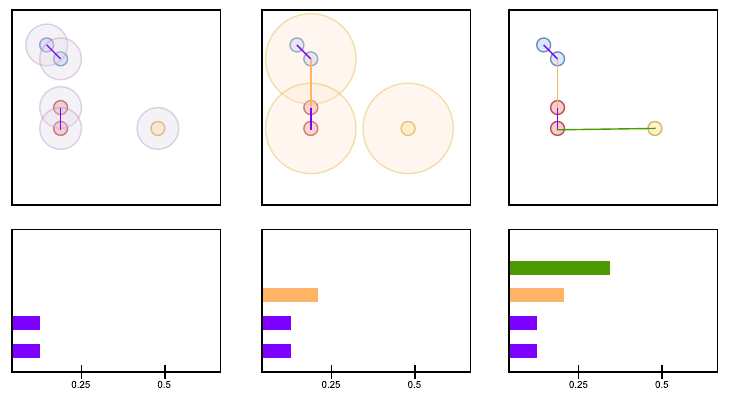} 
    \caption{$H_0$ barcode construction. As the threshold increases, the separate connected components merge, resulting in the death of topological features. The horizontal axis is a sequence of thresholds $\varepsilon$, and each
horizontal bar corresponds to a single feature. } \label{fig:barcodes}
\end{figure}

\paragraph{Persistent homology.} As the threshold $\varepsilon$ increases, new topological features (e.g., connected components, holes) can appear and disappear. The persistent homology tool tracks the dynamics of these topological features. Formally, the $k$-th persistent homology of $S$ is the pair of sets of vector spaces $\{H_k(S_i) \, | \, 0 \leq i \leq n\}$ and
maps $f_{ij}$, where $f_{ij}: H_k(S_i) \rightarrow H_k(S_j)$ is a map induced by the embedding $S_i \subseteq S_j$. Each persistent homology class in this sequence is ``born'' at some $S_i$
and ``dies'' at some
$S_j$ or never dies~\cite{SABarannikov1994}. This birth-death process of a basic set of independent topological features can be visualized as the set of intervals $[\varepsilon_\mathrm{birth}, \varepsilon_\mathrm{death}]$ called barcode (see Figure~\ref{fig:barcodes}). The features with $0$ lifespans are typically excluded.  The horizontal axis is a sequence of thresholds $\varepsilon$, and each
horizontal bar corresponds to a single feature. We begin with $|X| = m$ connected components (all
of them are ``born''), and as $\varepsilon$ increases, their pairs are merged (each merge corresponds to a ``death'' of a feature). The $0-$th barcode construction procedure is equivalent to Kruskal's algorithm for minimum spanning tree (MST), the bars in the barcode correspond to the edges in the MST of $X$~\cite{tulchinskii2023topological}.

\section{Properties of $\operatorname{MTop-Div}_G(R, P)$}\label{appendProof31}
\paragraph{Proof of Proposition~\ref{prop:stability_properties}.} The $0-$th Cross-Barcode coincides with the set of edges in the minimal spanning tree of the weighted graph $G$ with all the weights within the $P$-vertex subset equal zero. Excluding the zero-weight edges, this edge set coincides with the minimal spanning forest that attaches the vertex set $R$ to the $P$ vertices. \qed 

\paragraph{Other properties of MTop-Div for attention graphs.}
Here we present some other properties of our proposed $\operatorname{MTop-Div}_G(R, P)$.
\begin{proposition}\label{prop:basic_properties}
The following holds for any attention graph $G$ with vertex set $V_G$ and its complementary vertex subsets $P, R$,  where $P \cup R = V_G$ and $P \cap R = \varnothing$. 
\begin{itemize}
    \item The divergence itself is bounded by
    \begin{equation}
        0 \leq \operatorname{MTop-Div}_G(R, P) \leq |R|.
    \end{equation}
    \item \textbf{(Stability.)} If the weights of $G$ change by no more than $\varepsilon$, then the corresponding $\operatorname{MTop-Div}(R, P)$ changes by no more than $\delta = \varepsilon |R|$. 
    \item \textbf{(Exact sequence.)} For any $\alpha$, the following sequence of natural maps of homology groups is exact
    \begin{align*}
    (\mathbb{Z}/2\mathbb{Z})^{\lvert P\rvert} &\xrightarrow{r_2}H_0(VR_{\alpha} (G))\xrightarrow{r_1} \\ & \xrightarrow{r_1} H_0(VR_{\alpha} (G, w_{(R \cup P) / P})) \xrightarrow{r_0} 0.
    \end{align*}
    \item \textbf{(Connection with hallucinations.)} The normalized divergence value $\frac{1}{|R|}\operatorname{MTop-Div}(R,P) = 0$ iff the MSF attaches every response token to a prompt token by a subtree with attention weights $ = 1$.
    
\end{itemize}   
\end{proposition} 

\textbf{Proof of Proposition~\ref{prop:basic_properties}.} \\ 1. This property is immediately obtained from the properties of an attention map: all its weights lie between $0$ and $1$. \qed \\ 2. Denote by $\operatorname{MSF}(R, P)$ the minimum spanning forest attaching $R$ to $P$.
Note that we have properties~\ref{prop:basic_properties}, so
\begin{equation}
    \operatorname{MTop-Div}(R, P) = \sum_{e \in \operatorname{MSF}(R, P)} w(e).
\end{equation}
Therefore, we have to show that the weight of $\operatorname{MSF}(R, P)$ does not change significantly when all weights are changed by no more than $\varepsilon$. 

There are two possibilities: 1)  after a change, all MSF edges remain the same, or 2) some edges are replaced with other edges. In the first case, it is obvious that the total sum of edge weights changes by no more than $$\delta = \varepsilon \cdot \#\text{edges}(\operatorname{MSF}(R, P)) = \varepsilon \cdot |R|.$$ Consider the second case. Denote by $\operatorname{MSF}_{\mathrm{prev}}$ the original MSF, by $\operatorname{MSF}_{\mathrm{new}}$~---~the MSF after the change; let $w$ be the edge weight function before the change, $\hat{w}$~---~after the change. The following inequalities hold:
\begin{align}
    \hat{w}(\operatorname{MSF}_{\mathrm{new}}) &\leq \hat{w}(\operatorname{MSF}_{\mathrm{prev}});\label{eq9} \\ 
    w(\operatorname{MSF}_{\mathrm{prev}}) - \delta \leq \hat{w} &(\operatorname{MSF}_{\mathrm{prev}}) \leq \nonumber \\
    &\leq w(\operatorname{MSF}_{\mathrm{prev}}) + \delta; \label{eq10} \\
    w(\operatorname{MSF}_{\mathrm{new}}) - \delta \leq \hat{w} &(\operatorname{MSF}_{\mathrm{new}}) \leq \nonumber\\
   &\leq  w(\operatorname{MSF}_{\mathrm{new}}) + \delta; \label{eq11} \\
    w(\operatorname{MSF}_{\mathrm{new}}) &\geq w(\operatorname{MSF}_{\mathrm{prev}}).\label{eq12}
\end{align}
From~\eqref{eq9}-\eqref{eq10} follows that $\hat{w}(\operatorname{MSF}_{\mathrm{new}}) <  w(\operatorname{MSF}_{\mathrm{prev}}) + \delta$; from~\eqref{eq11}-\eqref{eq12} follows that $\hat{w}(\operatorname{MSF}_{\mathrm{new}}) \geq w(\operatorname{MSF}_{\mathrm{prev}}) - \delta$. \qed  \\
3. We have to check the definition of the exact sequence: $\text{Ker}(r_i) = \text{Im}(r_{i + 1})$. For a pair $r_0, \, r_1$, it is equivalent to the surjectivity of $r_1$. The $H_0$ homology group of a graph corresponds to the connected components of the graph. The set of edges $E_{(G,w)}^{\leq \alpha}=\{e\in E_G\vert w_e\leq\alpha\}$ is always a subset in the analogous set of the weighted graph $(G,w_{(R \cup P) / P})$ with all weight edges between $P$ vertices set to zero. Therefore, the map $r_1$ between their connected components is surjective. Similarly, the kernel of the map $r_1$ is spanned by the differences of two connected components, which are merged after adding some of the edges between $P$ vertices, and any such difference lies in the image of the map $r_2$. Also, any two vertices from $P$ belong to the same connected component in the graph $(G,w_{(R \cup P) / P}\leq\alpha)$, hence the image of $r_2$ is in the kernel of $r_1$. Therefore, the considered sequence is exact indeed. \qed 
4. Follows obviously from the MSF formula for $\operatorname{MTop-Div}(R, P)$ and attention map properties. \qed \\
\paragraph{Intuition behind MTop-Div properties and hallucination detection.} The stability property guarantees that similar attention patterns yield similar hallucination scores, making the metric's behavior consistent and predictable. The exact sequence property formalizes the geometric intuition behind our metric, which measures the strength of the response’s connection to the prompt through multiscale topological features of the attention graph. In the last property, we present an ``ideal'' case: if a model ``knows what to look at''~---~each token in the response attends to some token in the prompt with an attention weight equal to 1~---~$\operatorname{MTop-Div}$ would be equal to $0$, indicating zero uncertainty.

\section{Datasets}
\label{sec:appendix_datasets}

SQuAD~\cite{rajpurkar2016squad} and CoQA~\cite{reddy2019coqa} are widely used English question-answering benchmarks that have facilitated the development of hallucination detection datasets~\cite{kuhn2023semantic, manakul2023selfcheckgpt}. Similarly, XSum~\cite{narayan-etal-2018-dont}, a dataset of news articles with one-sentence summaries, is commonly employed in hallucination detection research for abstractive summarization~\cite{shi2024trusting, cao-etal-2022-hallucinated}. To assess LLM performance, we used GPT-4o to annotate responses to questions sourced from SQuAD, CoQA, and summarization tasks from XSum dataset.

\begin{table*}[hp]
\centering
\resizebox{\textwidth}{!}{\begin{tabular}{ll}
        \hline
        SQuAD  & CoQA \\
        \hline
        Given the context, answer the question in a brief but complete sentence. & Once upon a time, in a quiet village, there lived a kind old baker named Henry. \\
        Note that your answer should be strictly based on the given context. & He was known for his delicious bread and warm smile. One day, a traveler arrived, \\
        In case the context does not contain the necessary information to answer the question, & tired and hungry, Henry welcomed him with a fresh loaf.\\
        please reply with ``Unable to answer based on given context''. & \textit{Q:} What was Henry known for? \\
        \textit{Context:} & \textit{A:} Baking delicious bread. \\
        Once upon a time, in a quiet village, there lived a kind old baker named Henry. &  \textit{Q:} What else? \\
        He was known for his delicious bread and warm smile. One day, a traveler arrived, & \textit{A:} Warm smile. \\
        tired and hungry, and Henry welcomed him with a fresh loaf. & \textit{Q:} How did the traveler feel when he arrived? \\
        \textit{Question:} Who was known for baking delicious bread? & \textit{A:}  Tired and hungry. \\
        \textit{Answer:} & \textit{Q:} What did Henry give the traveler?\\
        \hline
        
    \end{tabular}}
\vspace{3mm}
\caption{Examples of prompts used during generation for CoQA and SQuAD (we add additional delimiter spaces and formatting not present in actual prompts for better readability). SQuAD contains instructions followed by context and questions. In CoQA, the prompt has only a contextual passage followed by a question-and-answer series, with the last question being the actual one. }
\label{tab:prompts}
\end{table*}

\begin{table*}[hp]
\centering
\resizebox{\textwidth}{!}{\begin{tabular}{l}
        \hline
        XSum \\
        \hline
       Please annotate potentially hallucinated model-generated summaries in the following settings. \\
    I will provide a reference text and a model-generated summary of this text. You will judge whether the given model-generated \\ summary contains hallucinations. 
    Answer ``Yes'' if the summary contains hallucinations,  ``No'' if it does not, and "N/A" if you cannot decide. \\ Do NOT give any extra explanations. \\
        \hline
    \end{tabular}}
\vspace{3mm}
\caption{The prompt used during generation for the XSum dataset (we add additional delimiter spaces and formatting not present in actual prompts for better readability).}
\label{tab:prompt_xsum}
\end{table*}

\begin{table*}
\centering
\scriptsize
\resizebox{\textwidth}{!}{
    \begin{tabular}{l}
        \hline \\
        You are an AI assistant specialized in detecting hallucinations in question-answering tasks. \\
        Your job is to analyze the given context, question, and generated answer to identify \\
        whether the answer contains any hallucinations. Examples: \\\\
        Example 1. \\
        \textit{Context}: \\
        The city of Paris is the capital of France. It is known for its iconic landmarks \\
        like the Eiffel Tower and Notre Dame Cathedral. \\
        The city is situated in the northern part of the country, near the Seine River. \\
        \textit{Question}: Is Paris the capital of Germany? \\
        \textit{Generated answer}: Yes, Paris is the capital of Germany. \\
        \textit{Hallucination}: Yes. \\\\
        Example 2. \\
        \textit{Context}: \\
        The city of Paris is the capital of France. \\
        It is known for its iconic landmarks like the Eiffel Tower and Notre Dame Cathedral. \\
        The city is situated in the northern part of the country, near the Seine River. \\
        \textit{Question}: Is Paris the capital of Germany? \\
        \textit{Generated answer}: No, Paris is not the capital of Germany. According to the context, \\ Paris is the capital of France. \\
        \textit{Hallucination}: No. \\\\             
        You should determine if the answer contains hallucinations according to the hallucination types above. \\
        If you cannot decide if the generated answer is a hallucination, write ``N/A.'' as the answer. \\
        The answer you give MUST be ONLY ``Yes.'', ``No.'' or ``N/A.''; do NOT give ANY explanation.\\
        \\\hline
    \end{tabular}
}
\vspace{3mm}
\caption{Example of annotation prompt passed to GPT-4o (we add additional delimiter spaces and formatting not present in actual prompts for better readability).}
\label{tab:annot_inst}
\end{table*}

\begin{table*}[!htp]
\centering
 \scriptsize
\resizebox{\textwidth}{!}{
    \begin{tabular}{llcccccc}
      \hline
      \multicolumn{2}{c}{Prompt number} & 1 & 2 & 3 & 4 & 5 & Average \\
      \hline
      \multirow{3}{*}{CoQA} & Accuracy ($\uparrow$) & 0.809 ± 0.017 & 0.861 ± 0.015 & 0.742 ± 0.003 & 0.795 ± 0.009 & 0.831 ± 0.025 & 0.808\\
      & Precision ($\uparrow$) & 0.849 ± 0.021 & 0.911 ± 0.007 & 0.771 ± 0.003 & 0.828 ± 0.011 & 0.860 ± 0.012 & 0.844\\
      & Recall ($\uparrow$) & 0.871 ± 0.004 & 0.877 ± 0.019 & 0.877 ± 0.013 & 0.877 ± 0.005 & 0.893 ± 0.027 & 0.879\\
      \hline
      \multirow{3}{*}{SQuAD} & Accuracy ($\uparrow$) & 0.831 ± 0.003 & 0.857 ± 0.018 & 0.857 ± 0.008 & 0.872 ± 0.003 & 0.854 ± 0.007 & 0.854 \\
      & Precision ($\uparrow$) & 0.813 ± 0.002 & 0.831 ± 0.028 & 0.845 ± 0.021 & 0.850 ± 0.011 & 0.847 ± 0.007 & 0.837 \\
      & Recall ($\uparrow$) & 0.796 ± 0.008 & 0.839 ± 0.010 & 0.823 ± 0.023 & 0.858 ± 0.018 & 0.813 ± 0.017 & 0.826 \\
      \hline
    \end{tabular}
    \newline
    \vspace*{0.4cm}
    \newline
}
\vspace{2mm}
\caption{Classification metrics of GPT-4o annotation for CoQA and SQuAD with human labels considered as true. The table shows metric scores for different prompt variants and the average score across all variants.}
\label{tab:human_vs_gpt}
\end{table*}

\subsection{Data Generation \& Annotation}

\paragraph{Generation.} We generate responses from a language model (LLM) for the considered datasets, employing different prompting strategies for each dataset while keeping these strategies consistent across models (see prompt examples in Table~\ref{tab:prompts}). For SQuAD and XSum, responses are generated using a zero-shot approach. In contrast, for CoQA, we generate queries in a few-shot manner without providing specific instructions, following~\cite{lin2024generating}: each sample consists of a passage and a series of question-answer pairs, culminating in a final question that the model is expected to answer. 

\paragraph{Annotation: automated vs human.} We treat hallucination detection as a binary classification problem; our target indicates whether a hallucination is present anywhere in the model's response. Two approaches to annotating model generations were considered: 1) automated annotation using an LLM (in our case, GPT-4o), and 2) manual annotation by human experts.

During the automated annotation process, we provide GPT-4o with an LLM's output, preceded by an instruction (prompt). In this prompt, GPT-4o is asked to determine whether the output contains hallucinations, and we expect a single-word response of either ``Yes'' or ``No.'' An example of such an instruction for the question answering task is shown in Table~\ref{tab:annot_inst}. 

For human annotation, we asked three team members with at least upper-intermediate English proficiency to independently annotate approximately 100 samples from each dataset. The human annotation consistency metrics are provided in Table~\ref{tab:consistency}. We selected samples for which all annotators reached consensus and treated these annotations as the ground-truth hallucination labels.

\begin{table}[]
\centering
    \begin{tabular}{lcc}
    \hline
Pair & Accuracy & Pearson-r
\\
\hline
1 vs 2 & 0.783 & 0.6 \\
1 vs 3 & 0.812 & 0.631 \\
2 vs 3 & 0.826 & 0.653 \\
\hline
Mean & 0.807 & 0.628 \\
Std & 0.022 & 0.027 \\
\hline
\end{tabular}
    \caption{Cross-annotator consistency metrics. }
    \label{tab:consistency}
\end{table}

To further evaluate GPT-4o, we conducted automatic annotation using several variations of prompts, each reformulating the task for GPT-4o, including zero-shot and few-shot versions. We then compared these annotations to the actual hallucination labels. The results, presented in Table~\ref{tab:human_vs_gpt}, demonstrate a consistent alignment between GPT-4o’s annotations and those made by humans, regardless of the specific prompt. This consistency confirms the robustness of our approach to the exact form of instruction. 

Based on these findings, we prefer automated annotation as a cost-effective and efficient alternative to human experts.

\paragraph{Annotation: general pipeline.} CoQA and SQuAD contain questions paired with ground-truth answers. To minimize false positives in labeling, we employed a two-step verification process:  
\begin{enumerate}
    \item Rouge-L scoring: we computed Rouge-L scores (using the \texttt{evaluate} library, v0.4.6) between the model’s response and the ground-truth answers.  
    \item Substring matching: we checked whether any ground-truth answer was a substring of the response.
\end{enumerate}
Responses with a Rouge-L score of 1 (exact match) were labeled as grounded. Those meeting both of the following criteria were flagged as potential hallucinations:  
\begin{itemize}
    \item Rouge-L score $\leq 0.3$ (following ~\cite{kuhn2023semantic}); 
    \item no ground-truth answer appears as a substring.
\end{itemize}
These candidate hallucinations were then reviewed by GPT-4o, and only confirmed cases were finally labeled as hallucinations.  

For XSum, where reference summaries are more complex than the ground truth answers in SQuAD/CoQA, we bypassed Rouge-L filtering and relied solely on GPT-4o for annotation.  

\begin{figure*}[t!]
    \centering
    \includegraphics[width=\textwidth]{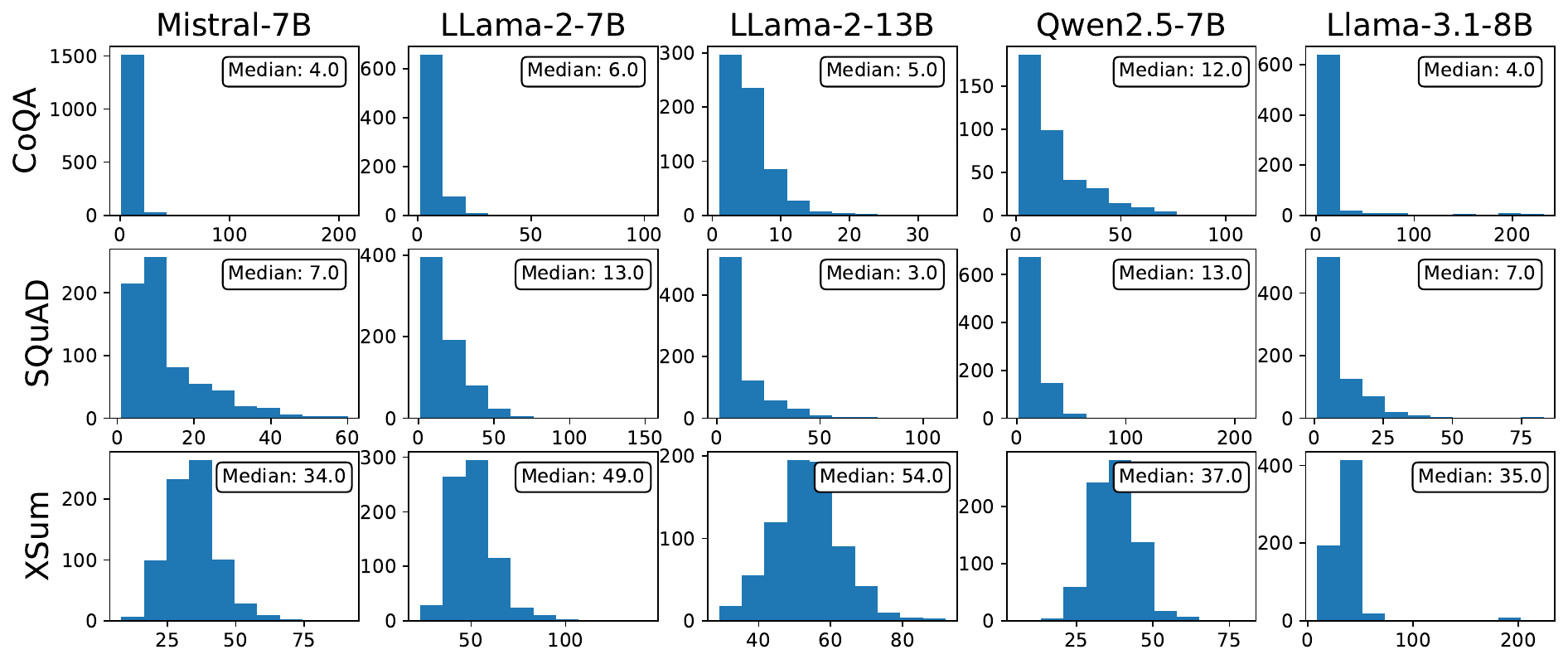} %
    \caption{The distributions of response lengths (in words).} \label{fig:stats}
\end{figure*}

Detailed statistics for each dataset are shown in Table \ref{tab:datasets_statistics}. 
The number of samples in the datasets varies across models, as we sought to maintain a balance between hallucinated and grounded responses, ensure sample cleanliness, and minimize mislabeling. The procedure outlined above selects a different number of objects in a sample depending on the quality of the model's responses. Additionally, we provide statistics on the response length distribution for each dataset we collected (Figure~\ref{fig:stats}).

\begin{table}
\centering
    \resizebox{\columnwidth}{!}{\begin{tabular}{lcccccc}
    \hline
     \multirow{2}{*}{Model} & \multicolumn{2}{c}{CoQA} & \multicolumn{2}{c}{SQuAD} &  \multicolumn{2}{c}{XSum} \\
    
     & Hal. & Grounded & Hal. & Grounded & Hal. & Grounded \\ 
    \hline
    Mistral-7B & 776 & 776 & 311 & 389  & 301 & 448 \\ 
    LLaMA-2-7B & 375 & 375 & 440 & 258  & 239 & 507 \\ 
    LLaMA-2-13B & 279 & 384 & 314 & 436  & 208  & 522 \\ 
    LLaMA-3.1-8B & 356 & 350 & 350 & 400 & 243 & 407 \\ 
    Qwen2.5-7B & 171 & 218 & 423 & 423 & 194 & 556 \\ 
     
    \hline
    \end{tabular}}
\caption{Datasets statistics. Number of hallucinated and grounded samples of each model. }
\label{tab:datasets_statistics}
\end{table}

\section{Implementation Details}\label{sec:appendix_implementation}
In this section, we describe the key implementation choices.

\begin{itemize}

\item For EigenScore, we used the last token representation to embed sentences, as suggested in~\cite{chen2024inside}. We took outputs from the 16th layer, since middle layers were shown to contain the most factual information~\cite{sky2024androids, azaria-mitchell-2023-internal}.

\item For SelfCheckGPT, we used its NLI-based version.

\item For LLM-Check, we considered its white-box attention score modification, as it works in a setting similar to ours.

\item The topological divergences were calculated using \texttt{ripser} library~\cite{ctralie2018ripser}, MIT license.
\item For the RAGTruth dataset, the model settings were aligned with the original paper.
\end{itemize}

All experiments were carried out on an NVIDIA L40. 
 
\section{Use of scientific artifacts \& AI assistants}

CoQA contains passages from seven domains under the following licenses:  Literature and Wikipedia passages are shared under CC BY-SA 4.0 license; Children's stories are collected from MCTest, which comes with the MSR-LA license; Middle/High school exam passages are collected from RACE which comes with its own license; News passages are collected from the DeepMind CNN dataset which comes with Apache license. The SQuAD dataset is licensed under CC BY-SA 4.0. The RAGTruth dataset comes under the MIT license. The XSum dataset comes under the MIT license.

We used all the artifacts as it was intended by the corresponding licenses. No personal information or offensive content is contained in the considered datasets. 

The original text of this paper was spell- and grammar-checked and slightly smoothed using Grammarly.
\section{Potential risks}

\begin{enumerate}
    \item Ethical risks from deployment: overconfidence in TOHA’s scores could lead to unchecked LLM outputs in high-stakes scenarios (e.g., healthcare). TOHA should be framed as a ``warning system'' rather than a definitive filter, and advocate for human review.
    \item Attention manipulation attacks: adversarial prompts could artificially alter attention patterns, evading detection. 
\end{enumerate}



\end{document}